\pdfoutput=1
\documentclass[11pt,a4paper]{article}
\usepackage{acl}
\usepackage{times}
\usepackage{latexsym}

\usepackage{booktabs}
\usepackage{chngcntr}
\usepackage{makecell}

\usepackage{multirow}

\usepackage{microtype}
\usepackage{mathtools}

\newcommand{\lang}[1]{\textsc{\MakeLowercase{#1}}}

\usepackage[noabbrev,capitalize,nameinlink]{cleveref}
\crefname{section}{\S}{\S\S}
\Crefname{section}{\S}{\S\S}    
\crefformat{section}{#2\S#1#3}  
\Crefformat{section}{#2\S#1#3}
\crefrangeformat{section}{\S\S#3#1#4--#5#2#6}
\Crefrangeformat{section}{\S\S#3#1#4--#5#2#6}
\crefformat{section}{#2\S#1#3}  
\crefrangeformat{section}{\S\S#3#1#4--#5#2#6}

\def\Snospace~{\S{}}

\newcommand{\taskabbrev}{TrUMP}
\renewcommand{\taskabbrev}{\textit{t}UMPC}
\newcommand*\samethanks[1][\value{footnote}]{\footnotemark[#1] }

\DeclareMathOperator*{\argmax}{argmax}
\title{Morphological Processing of Low-Resource Languages:\\Where We Are and What's Next}

\author{
Adam Wiemerslage${ }^{\sharp}$\thanks{*The first two authors contributed equally.} \quad
Miikka Silfverberg${ }^{\flat}$\samethanks \quad
Changbing Yang${ }^{\flat}$ \quad
Arya D. McCarthy${ }^{\natural}$ \\
\textbf{Garrett Nicolai${ }^{\flat}$  \quad
Eliana Colunga${ }^{\sharp}$ \quad
Katharina Kann${ }^{\sharp}$} \\
${ }^{\sharp}$University of Colorado Boulder \quad
${ }^{\flat}$University of British Columbia \\
${ }^{\natural}$Johns Hopkins University \quad}

\date{}

\begin{document}
\maketitle

\begin{abstract}
Automatic morphological processing can aid downstream natural language processing applications, especially for low-resource languages, and assist language documentation efforts for endangered languages. Having long been multilingual, the field of computational morphology is increasingly moving towards approaches suitable for languages with minimal or no annotated resources. 
First, we survey recent developments in computational morphology with a focus on low-resource languages. Second, we argue that the field is ready to tackle the logical next challenge: understanding a language’s morphology from raw text alone. We perform an empirical study on a truly unsupervised version of the paradigm completion task and show that, while existing state-of-the-art models bridged by two newly proposed models we devise perform reasonably, there is still much room for improvement. The stakes are high: solving this task will increase the language coverage of morphological resources by a number of magnitudes.
\end{abstract}

\section{Introduction}
Automatic morphological processing tools have the potential to drastically speed up language documentation \cite{moeller-etal-2020-igt2p} and thereby help combat the language endangerment crisis \citep{austin_sallabank_2011}. Explicit morphological information also benefits myriad  NLP tasks, such as parsing \cite{hohensee-bender-2012-getting,seeker-cetinoglu-2015-graph}, language modeling \cite{blevins-zettlemoyer-2019-better,10.1162/tacl_a_00365,hofmann-etal-2021-superbizarre}, and machine translation \cite{dyer-etal-2008-generalizing,tamchyna-etal-2017-modeling}. 

For low-resource languages, valuable morphological resources are typically small or non-existent. 
Of late, the field of computational morphology has increased its efforts to extend the coverage of multilingual morphological resources \cite{kirov-etal-2016-large,kirov-etal-2018-unimorph,mccarthy-etal-2020-unimorph,metheniti-neumann-2020-wikinflection}. Simultaneously, there has been a revival of minimally supervised and unsupervised models for morphological tasks, such as segmentation \cite{eskander-etal-2019-unsupervised}, inflection \cite{kann-etal-2017-one}, and lemmatization \cite{bergmanis-goldwater-2019-data}. 
Given the speed of recent developments, it is important to reflect on where we are as a field and what future challenges lie ahead. 

To this end, we survey recent computational morphology: we review existing multilingual resources (\autoref{sec:resources}) and tasks and systems (\autoref{sec:tasks}), with a focus on low-resource languages. Given recent developments in unsupervised segmentation, low-resource morphological inflection, and unsupervised morphological paradigm completion  \cite{jin-etal-2020-unsupervised,erdmann-etal-2020-paradigm}---which we argue is not fully unsupervised---we believe the community is poised for the next logical step: inferring a language’s morphology purely from raw text. 

In \Cref{sec:whats-next}, we formalize a new task: \textit{truly unsupervised morphological paradigm completion} (\taskabbrev). We then introduce a pipeline with two novel components (\autoref{sec:slot-alignment}): one model for aligning paradigm slots across lexemes and another for predicting the slots of observed forms. With these, we assess several state-of-the-art models and the influence of different types of unlabeled corpora within the framework of \taskabbrev. While existing methods leave room for improvement, they perform reasonably enough to support our argument that inferring a language’s morphology from raw text is within reach and worthy of community efforts. 

To summarize, we present the following contributions: (i) 
a survey of tasks and systems in computational morphology with a focus on low-resource languages;
(ii) models for the tasks of paradigm slot alignment and slot prediction, (iii) a formalization of the task of truly unsupervised morphological paradigm completion and (iv) an evaluation of state-of-the-art approaches and different corpora within the framework of this task.  Our code and data are publicly available.\footnote{\url{https://github.com/Adamits/tUMPC}}

\section{Morphological Resources}
\label{sec:resources}
Manually created resources are necessary for developing and evaluating NLP systems. They also serve as a basis for research questions in a multilingual context \cite{pimentel-etal-2019-meaning,wu-etal-2019-morphological}.\footnote{This approach has been criticized by \citet{malouf-etal-2020-lexical} due to incompleteness and quality of existing resources.}
Below, we review the two largest active multilingual resources for morphology and a number of language-specific resources.

\paragraph{Background and Notation}
The canonical form of a word is called its \emph{lemma}, and the set of all surface forms of a lemma is referred to as that lemma's \emph{paradigm}. As is common, we formally write the paradigm of a lemma~$\ell$ as:
\begin{equation}
    \pi(\ell) = \left\langle f(\ell, \vec{t}_\gamma)\right\rangle_{\gamma \in \Gamma(\ell)},
\end{equation}
with $f : \Sigma^* \times \mathcal{T} \to \Sigma^*$ defining a mapping from a tuple consisting of the lemma and a vector $\vec{t}_\gamma \in \mathcal{T}$ of morphological features to the corresponding inflected form. $\Sigma$ is an alphabet of discrete symbols: the characters used in the language of lemma~$\ell$. $\Gamma(\ell)$ is the set of slots in $\ell$'s paradigm. 

\paragraph{UniMorph}
The UniMorph project \cite{10.1007/978-3-319-23980-4_5,sylak-glassman-etal-2015-language,kirov-etal-2016-large} is a database of triples organized into paradigms, where each triple represents a word as its lemma $\ell$, morpho-syntactic description $\vec{t}_\gamma$, and surface form $f(\ell, \vec{t}_\gamma)$.  An English example triple is: 
\begin{equation}
    \textrm{mutate}	\qquad \textrm{mutates}	\qquad \textrm{V;3;SG;PRS}\nonumber
\end{equation}
This structure provides training data for inflection generation, lemmatization, or paradigm completion. The most recent version of UniMorph \cite{mccarthy-etal-2020-unimorph}  includes 118 languages and 14.8 million triples, with more languages under development.
As it is semi-automatically created,
issues have been noted---particularly, it is a convenience sample across languages \cite{gorman-etal-2019-weird,malouf-etal-2020-lexical}. 
Still, related efforts validate themselves using UniMorph, including \citet{metheniti-neumann-2020-wikinflection}---another Wiktionary-derived resource for morphology. Wikinflection captures segmentation information (\cref{sec:segmentation}) from Wiktionary templates, though the authors note some limits in the morphological tags that are extracted to accompany these.

\paragraph{Universal Dependencies}
Whereas UniMorph contains type-level annotations, the Universal Dependencies project (UD) is a resource of token-level annotations. As of writing, the latest release \citep[v2.8;][]{11234/1-3683} spans 114 languages, typically semi-automatically extracted from existing corpora, sometimes with less comprehensive annotations \citep{malaviya-etal-2018-neural}. The structure is useful for morphological tagging (\autoref{sec:morph-tagging}) at the sentence level \citep{goldman-tsarfaty-2021-well}, and several languages have parallel text, enabling evaluation of projection-based approaches for morphology induction, parsing, and other tasks \citep{yarowsky-etal-2001-inducing,rasooli-collins-2017-cross}.

\paragraph{Mapping between UniMorph and Universal Dependencies}
The UD2 morphological annotations borrow several features from UniMorph.\footnote{\url{http://universaldependencies.org/v2/features.html\#comparison-with-unimorph}} Consequently, there is great harmony between the two schemas. A deterministic mapping
\citep{mccarthy-etal-2018-marrying} has shown the synergy; for instance, \citet{bergmanis-goldwater-2019-data} augment a contextual tagger with UniMorph inflection tables.


\paragraph{Language-Specific Resources} Throughout the years, many language-specific morphological resources have been created. These include corpora and treebanks like the morphologically annotated corpus for Emirati Arabic by \newcite{khalifa2018morphologically}. Resources also come in the form of morphological databases, such as CELEX for Dutch, English and German  \cite{baayen1996celex}, or morphological analyzers, such as the Paraguayan Guaraní analyzer presented by \newcite{zueva-etal-2020-finite}. 

Creation of morphological resources is an ongoing effort which in recent years has increasingly focused on low-resource languages. Several conferences and workshops like LREC \cite{lrec-2020-language}, SIGMORPHON \cite{sigmorphon-2021-sigmorphon}, ComputEL \cite{computel-2021-use}, AmericasNLP \cite{mager2021findings}, PYLO \cite{ws-2018-modeling-polysynthetic} and FSMNLP \cite{maletti2011proceedings} have presented and continue to present language-specific tools and datasets for computational morphology.

\section{Where We Are: Tasks and Systems} \label{sec:tasks}
\subsection{Morphological Tagging} \label{sec:morph-tagging}
Morphological tagging is a sequence-labeling task similar to part-of-speech (POS) tagging. As a token-level task, it considers words in context. 
Given a sentence, it consists of assigning 
to each word $f(\ell, \vec{t}_\gamma)$ a morphosyntactic description (MSD), i.e., a tag representing the morphological features $\vec{t}_\gamma$ it expresses. For instance, in the sentence \textit{The virus mutates}, the word
\textit{mutates} would be assigned the tag \textsc{\MakeLowercase{V;3;SG;PRS}}.
Morphological tagging was featured in the SIGMORPHON 2019 shared task \cite{mccarthy-etal-2019-sigmorphon}. 

\paragraph{Systems }
A leading non-neural morphological tagger is MARMOT \cite{mueller-etal-2013-efficient}, a higher-order conditional random field \citep[CRF; ][]{lafferty2001conditional} tagger. Of late, LSTM \cite{hochreiter1997long} and Transformer \cite{vaswani2017attention} models have been used for tagging \cite{heigold2016neural, heigold-etal-2017-extensive,nguyen-etal-2021-trankit}.

For low-resource languages, both projection-based approaches \cite{buys-botha-2016-cross} and cross-lingual transfer approaches via multitask training \cite{cotterell-heigold-2017-cross} have been developed. 16 systems were submitted to the SIGMORPHON 2019 shared task\footnote{The task is concerned with joint lemmatization and tagging, but systems can be used for separate tagging as well.} \cite{mccarthy-etal-2019-sigmorphon}, which featured 66 languages. The winning team \cite{kondratyuk-2019-cross} built a tagger based on multilingual BERT \cite{devlin-etal-2019-bert}, thus employing cross-lingual transfer; for other systems, we refer the reader to the shared task overview. The largest multilingual morphological tagging effort to date is that by \citet{nicolai-etal-2020-fine} who build morphological analyzers for 1108 languages using projection from a high-resource to a low-resource language via the aligned text in the JHU Bible Corpus \cite{mccarthy-etal-2020-johns}.

\subsection{Morphological Segmentation} \label{sec:segmentation}
The goal of morphological segmentation \citep{goldsmith2010segmentation} is to split words into their smallest meaning-bearing units: morphemes. 
We discuss both surface and canonical segmentation here.

\subsubsection{Surface Segmentation}
During surface segmentation, a word is split into morphemes in a way such that the concatenation of all parts exactly results in the original word. An example (with ``*" marking boundaries) is:
\begin{equation}
    \textrm{mutates} \rightarrow \textrm{mutate * s}\nonumber
\end{equation}
Surface segmentation was the focus of the Morpho Challenge from 2005 to 2010 \cite{kurimo-etal-2010-morpho}. The competition featured datasets in Finnish, Turkish, German, English, and Arabic.
Additionally, segmentation was a track (alongside morphological analysis and generation) of LowResourceEval-2019 \cite{klyachko2020lowresourceeval}, a shared task which featured four low-resource languages from Russia. The shared task overview lists morphological resources for other Russian languages.

\paragraph{Systems } Many approaches to this task are unsupervised. \citet{harris1970morpheme} identifies morpheme boundaries in English based on the frequency of characters at the end of a word. LINGUISTICA \cite{goldsmith-2001-unsupervised} finds sets of stems and suffixes that represent the minimum description length of the data. MORFESSOR \cite{creutz-lagus-2002-unsupervised} introduces a family of probabilistic models for identifying morphemes, which have seen wide use, including variations of the original model \cite{virpioja2009unsupervised,smit2014morfessor}. \citet{lignos2009rule} learn rewrite rules that can explain many types in the corpus. \citet{poon2009unsupervised} apply a CRF to unsupervised segmentation by learning parameters with contrastive estimation \cite{smith2005contrastive}. Incorporating semantic similarity between related words that form "chains" has also been shown to be effective \cite{narasimhan2015unsupervised}.
\citet{monson2007paramor} propose a segmentation algorithm that exposes the properties of partial morphological paradigms in order to learn segments.
\citet{xu-etal-2018-unsupervised} iteratively refine segments according to their distribution across paradigms. They filter unreliable paradigms with statistically reliable ones, and induce segments with the proposed partial paradigms. Both systems can only model suffix concatenation. \citet{xu-etal-2020-modeling} follow a similar strategy, but incorporate language typology, expanding beyond suffixes, and outperform \citet{xu-etal-2018-unsupervised}. MorphAGram \cite{eskander-etal-2020-morphagram} is a publicly available tool for unsupervised segmentation based on adaptor grammars \cite{johnson2007adaptor}.

Supervised \cite{creutz2005inducing,ruokolainen-etal-2013-supervised,cotterell-etal-2015-labeled} and semi-supervised systems \cite{ruokolainen-etal-2014-painless} also exist. Non-neural systems are often based on CRFs. \citet{ruokolainen-etal-2013-supervised} focus explicitly on low-resource settings and perform experiments on Arabic, English, Hebrew, Finnish, and Turkish with training set sizes as small as 100 instances.

Neural models have also been applied to surface segmentation: \citet{wang2016morphological} obtain strong results with window LSTM neural networks in the high-resource setting, \citet{seker-tsarfaty-2020-pointer} introduce a pointer network \cite{vinyals2015pointer} for segmentation and tagging, and \citet{micher-2017-improving} propose a segmental RNN \cite{kong2015segmental} for segmentation and tagging of Inuktitut. \citet{kann-etal-2018-fortification} explore LSTM-based sequence-to-sequence (seq2seq) models for segmentation in combination with data augmentation, multitask and multilingual training; they evaluate on  datasets they introduce for four low-resource Mexican languages. \citet{eskander-etal-2019-unsupervised} apply an \textit{unsupervised} approach based on adaptor grammars to the same languages; it outperforms supervised methods in some cases. \citet{sorokin-2019-convolutional} show that CNNs outperform RNN-based models on that data as well as on North S\'{a}mi \cite{gronroos-etal-2019-north}. 

Additional contributions have been made by \citet{yarowsky-wicentowski-2000-minimally}, \citet{schone-jurafsky-2001-knowledge}, and \citet{Clark2001LearningMW}. Linguistically informed approaches show demonstrable value compared to approaches like BPE; see \citet{DBLP:journals/nle/Church20a} and \citet{hofmann-etal-2021-superbizarre}.
Still, not all morphological phenomena are suited for a segmentation-based analysis, as in fusional morphology that sometimes leaves ambiguity as to where a morpheme boundary lies; indeed in some cases there is no consensus among linguists as to the proper segmentation of a word. Therefore, (especially surface) segmentation is not necessarily meaningful for all languages.


\subsubsection{Canonical Segmentation}
Canonical segmentation 
is more complex:
its aim is to jointly split a word into morphemes and to undo the orthographic changes which have occurred during word formation. As a result, each word is segmented into its \textit{canonical} morphemes.
While often not being modeled this way in practice, the task can be seen as the following two-step process:
\begin{equation}
    \textrm{manic} \rightarrow \textrm{maniaic} \rightarrow \textrm{mania * ic}\nonumber
\end{equation}

\paragraph{Systems } The state-of-the-art pre-neural system is the CRF-based model by \citet{cotterell-etal-2016-joint}, which is jointly trained on segmentation and restoration of orthographic changes. The unsupervised system of \citet{bergmanis2017segmentation} builds upon MorphoChains \cite{narasimhan2015unsupervised}. Neural models are typically based on seq2seq architectures: \citet{kann-etal-2016-neural} use a seq2seq GRU and a feature-based reranker. Like \citet{cotterell-etal-2016-joint}, they evaluate on German, English, and Indonesian. \citet{ruzsics-samardzic-2017-neural} use a similar system, but add a language model over canonical segments and do not require external resources. In addition to German, English, and Indonesian, they evaluate on Chintang, a truly low-resource language spoken in Nepal. \citet{wang2019neural} use a character-level seq2seq model for (surface and) canonical segmentation in Mongolian. \citet{mager-etal-2020-tackling} show the benefit of copy mechanisms and introduce datasets for two low-resource Mexican languages. 
\citet{moeng2021canonical} show that Transformers outperform RNNs for canonical segmentation in four Nguni languages. 

\subsection{Lemmatization, Inflection, Reinflection}
Inflection and reinflection have recently gained popularity in computational morphology by being featured in yearly SIGMORPHON shared tasks \cite{cotterell2016sigmorphon}. They are concerned with generating inflected forms $f(\ell, \vec{t}_\gamma)$ of a lemma $\ell$; the target inflected form can be specified in different ways, depending on the exact task formulation. While the terms \textit{inflection} and \textit{reinflection} are sometimes used synonymously in the literature, \textit{inflection} refers to generating a word form from a given \textit{lemma}, while \textit{reinflection} refers to generation from an \textit{arbitrary} given form in the paradigm.
Lemmatization is a special case of reinflection: instead of generating an indicated inflected form, a lemma is produced. As the target form is implicitly determined by the task definition, lemmatization generally does not require tags to indicate which form to generate. 


\subsubsection{Type-level Versions} Most commonly, lemmatization, inflection and reinflection are type-level tasks. The input consists of an input form together with the target MSD (which can be omitted for lemmatization). The output is the corresponding inflected form, for instance: 
\begin{equation}
    \textrm{mutated}	~ \textrm{V;3;SG;PRS} \rightarrow \textrm{mutates}\nonumber
\end{equation}
The version of reinflection featured in the SIGMORPHON 2016 shared task also provides the MSD of the \textit{source form}, but performance improvements are usually minor \cite{cotterell-etal-2016-sigmorphon}.

\paragraph{Systems } Pre-neural systems for the task include those by \citet{durrett-denero-2013-supervised} and \citet{nicolai-etal-2015-inflection}.  These systems align lemmas and inflections before extracting character-level transductions for training CRF-inspired models.  \citet{faruqui-etal-2016-morphological} propose the first neural model for morphological inflection, an RNN seq2seq model, but fail to outperform prior approaches on some of the datasets they evaluate on. The breakthrough for neural models was the SIGMORPHON 2016 shared task \cite{cotterell-etal-2016-sigmorphon}, with about one third of the systems being neural: the winning system \cite{kann-schutze-2016-med,kann-schutze-2016-single} used multitask training by encoding MSDs together with the character sequence of the source word. This approach has now become the standard for the task, and while a multilingual version of the model by \citet{kann-schutze-2016-med} was submitted to the SIGMORPHON 2021 shared task 
\cite{pimentel-ryskina-etal-2021-sigmorphon,szolnok-barta-lakatos-etal-2021-bme}, the same multitask approach has since been used with other seq2seq models such as Transformers \cite{wu-etal-2021-applying}. Ensembles have been shown to improve performance for inflection \cite{kann-schutze-2016-med} and have been systematically studied for the task by \citet{kylliainen-silfverberg-2019-ensembles}.

The SIGMORPHON shared tasks on morphological inflection have focused increasingly on low-resource settings. Seq2seq models with hard monotonic attention \cite{aharoni-goldberg-2017-morphological}, a copy mechanism \cite{sharma-etal-2018-iit,singer-kann-2020-nyu}, or both \cite{makarov-etal-2017-align,makarov-clematide-2018-imitation,makarov-clematide-2018-neural} obtain great results for training sets as small as 100 examples. 
Cross-lingual transfer via multitask training 
was proposed by \citet{kann-etal-2017-one} for GRU seq2seq models and  has later been used with other architectures, e.g.,\ in the  SIGMORPHON 2019 shared task on cross-lingual transfer \cite{mccarthy-etal-2019-sigmorphon}.

Another approach suitable for low-resource languages is data augmentation. For morphological inflection, this was suggested by several contemporaneous works \cite{kann-schutze-2017-unlabeled,bergmanis-etal-2017-training,silfverberg-etal-2017-data}. In the following years, other augmentation strategies have been developed \cite{anastasopoulos-neubig-2019-pushing}. The success of data augmentation is mixed, as it is largely dependent on the architecture (\textit{Does it have to learn how to copy or is there a copy mechanism?}) as well as on the quality of the original data, which influences the quality of artificially generated examples.

\subsubsection{Token-level Versions} The token-level version of the task is often referred to as lemmatization or inflection \textit{in context}. Here the information about which form to generate is explicitly given via a sentence context in which the target word should be embedded, e.g.:
\begin{equation}
    \textrm{mutate -- }	~ \textrm{The virus [MASK].} \rightarrow \textrm{mutates}\nonumber
\end{equation}
A drawback of this formulation is that typically many different inflected forms are possible within the same context: in the given example, \textit{mutates} is the gold solution, but \textit{mutated} would be equally grammatical. To overcome this, multiple gold solutions can be provided \citep{cotterell-etal-2018-conll}. It might be impossible to unambiguously define the target form for some languages if the speaker's intention is unknown.

\paragraph{Systems } Lemmatization in context is arguably easier than inflection or reinflection, as the target form for generation is implicitly defined. Neural models for inflection are seq2seq architectures: \citet{bergmanis-goldwater-2018-context} propose Lematus, a character-level LSTM, which they later extend to the low-resource setting by training on labeled data in combination with raw text \cite{bergmanis-goldwater-2019-data}.
They explore data settings as small as 1k types each from UD and UniMorph. \citet{zalmout-habash-2020-utilizing} use a similar architecture to Lematus but add subword features. \citet{malaviya-etal-2019-simple} present a joint model for 
tagging and lemmatization and show that joint training benefits low-resource languages. They evaluate on 20 languages, using data from UD. The best lemmatizer in the SIGMORPHON 2019 shared task \cite{mccarthy-etal-2019-sigmorphon}, UDPipe \citep{straka-etal-2019-udpipe}, is based on BERT \cite{devlin-etal-2019-bert}.

\textit{Inflection} in context can be tackled by neural seq2seq models too. Models typically either see a context window around the target word \cite{makarov-clematide-2018-uzh,kann-etal-2018-nyu,acs-2018-bme} and then are optionally trained via multitask training \cite{kementchedjhieva-etal-2018-copenhagen} or predict the MSD of the form to generate as a first step \cite{liu-etal-2018-morphological}. \citet{kementchedjhieva-etal-2018-copenhagen} show that a multilingual model can aid low-resource languages via cross-lingual transfer.

\subsection{Paradigm Completion}
The paradigm cell filling problem \citep{ackerman2009parts} -- also called supervised paradigm completion \citep{cotterell-etal-2017-conll} -- is yet another inflection task, but differs from the above ones in that the inflected forms for \textit{all} slots $\Gamma(\ell)$ of lemma $\ell$'s paradigm need to be generated and that the input can consist of \textit{one or more} forms.


\paragraph{Systems } Many approaches for the paradigm cell filling problem are effectively systems for morphological reinflection and generate all forms of a paradigm individually and from a single input form, e.g., \citet{silfverberg-etal-2017-data,silfverberg-hulden-2018-encoder,moeller-etal-2020-igt2p}. \citet{kann-etal-2017-neural} propose a model for multi-source inflection, showing that multiple available forms per paradigm can be beneficial for generation, but do not evaluate on paradigm completion.
Two notable exceptions which design approaches explicitly for the paradigm cell filling problem are \citet{cotterell-etal-2017-neural} and \citet{kann-schutze-2018-neural}. \citet{cotterell-etal-2017-neural} rely on the notion of principal parts \cite{finkel2007principal} to jointly generate all forms in the paradigm. \citet{kann-schutze-2018-neural} use a transductive training approach, fine-tuning on a paradigm's \textit{input} forms before generating missing target forms. The latter shows good performance for training sets with as few as 10 paradigms. 

\subsection{Paradigm Clustering\label{sec:paradigm_clustering}}
Paradigm clustering can be seen as a first step towards the unsupervised analysis of a language's morphology and is typically part of pipelines for unsupervised paradigm completion (\autoref{subsec:upc}). The goal of paradigm clustering is to group all types in a corpus into
 (partial) morphological paradigms. 
For example, the input \textit{The}, \textit{virus}, \textit{mutates}, \textit{after}, \textit{it}, \textit{has}, \textit{mutated} should result in the paradigm cluster (\textit{mutates}, \textit{mutated}) and 5 singleton clusters.
Systems for the task can be evaluated using best-match F1 \citep[BMF1;][]{wiemerslage-etal-2021-findings}.

\paragraph{Systems \label{sec:paradigm_clustering_sys}} Perhaps the seminal work in distributionally-based paradigm clustering is the work of \citet{yarowsky-wicentowski-2000-minimally}. Their work predates embedding-based approaches while leveraging both distributional features of context and relative frequency, along with early statistical models of inflection-to-lemma string transduction. For instance, the work succeeds in identifying that the past tense of ‘sing’ is not ‘singed’ but ‘sang’, based on both the distributional signatures of music vs.\ fire terms in context, as well as the distribution of observed tense {\it frequency ratios}, where the regular {\it sing:singed} pairing can also be rejected given its frequency ratio is several standard deviations off of expectation, while the irregular {\it sing:sang} pairing occurs at nearly exactly the ratio expected.  While contextual information has been incorporated in follow-up works \citep{schone-jurafsky-2001-knowledge} and in recent approaches by means of word embeddings, we do not see much follow-on use of the frequency ratio features, which remain ripe for disambiguation of paradigm members.  

Segmentation approaches like \citet{goldsmith-2001-unsupervised}, developed to segment words into stems and affixes, can also be used to induce paradigm clusters. \citet{chan-2006-learning} formalizes the notion of a probabilistic paradigm --- modeling conditional probabilities of suffixes given paradigms and paradigms given stems. 
However, they that a segmentation is given, and only model regular morphology for unambiguous words, or those with a known POS. 
Some segmentation algorithms induce paradigms as a byproduct, as in \citet{monson2007paramor}, \citet{xu-etal-2018-unsupervised} and  \citet{xu-etal-2020-modeling}. These can also be employed as paradigm clustering systems.

Several systems have been proposed for the SIGMORPHON 2021 shared task \cite{wiemerslage-etal-2021-findings}. The best performing system \cite{mccurdy-etal-2021-adaptor} segments input types with MorphAGram \cite{eskander-etal-2020-morphagram}, then groups the resulting stems into paradigm clusters. \citet{yang-etal-2021-unsupervised} learn frequent transformation rules and cluster types together that result from rule application.

\subsection{Unsupervised Paradigm Completion}
\label{subsec:upc}
Due to the recent progress on supervised morphological tasks, unsupervised paradigm completion (UMPC; or the \textit{paradigm discovery problem} \cite{elsner2019modeling}) has recently (re)emerged as a promising way to automatically extend morphological resources such as UniMorph to more low-resource languages. Similar to the supervised version of the task, the goal is to generate the inflected forms corresponding to \textit{all} slots $\Gamma(\ell)$ of lemma $\ell$'s paradigm. However, no morphological annotations are given during training. Two independent
works propose similar unsupervised paradigm completion setups. In \citet{jin-etal-2020-unsupervised}, the basis of the SIGMORPHON 2020 shared task  \cite{kann-etal-2020-sigmorphon}, the input consists of 1) a corpus in a low-resource language and 2) a list of lemmas from one POS in that language.  In \citet{erdmann-etal-2020-paradigm}, the inputs are 1) a corpus and 2) a list of word forms belonging to a single POS. For both, the expected output is the paradigms for the words in the provided list.

As systems are trained without supervision, they 
cannot output human-readable MSDs and, instead, 
assign uninterpretable slot identifiers to generated forms. Thus, evaluation against gold standard data from UniMorph is non-trivial. \citet{jin-etal-2020-unsupervised} propose to evaluate systems via best-match accuracy (BMAcc):
the best accuracy among all mappings from pseudo tags to paradigm slots.

\paragraph{Systems } State-of-the-art systems for paradigm completion follow a pipeline approach similar to that by \citet{jin-etal-2020-unsupervised}: 1) based on the given input forms, they detect transformations which happen during inflection (and sometimes new lemmas), 2) the paradigm structure is detected based on the transformations, and 3) an inflection model is trained to generate missing surface forms. \citet{jin-etal-2020-unsupervised} employ the inflection model by \citet{makarov-clematide-2018-imitation}, while \citet{mager-kann-2020-ims} use the LSTM pointer-generator model from \citet{sharma-etal-2018-iit}, and \citet{singer-kann-2020-nyu} implement a Transformer-based pointer-generator model. The performance across languages is mixed \cite{kann-etal-2020-sigmorphon}. 

\paragraph{Is the Task Truly Unsupervised?}
Existing versions of the unsupervised paradigm completion task make small concessions to supervision requirements by providing lists of lemmas or surface forms from a single POS.  This simplifies a difficult task, but also makes it less realistic.  From the point of view of data availability, this method is not language-agnostic, as many languages do not have the required documentation: 
many of the world's languages have fuzzy POS definitions, and no annotated POS corpora.  From a language learning perspective, existing methods are closer to L2 than to L1 learning. 

Under this framing, UMPC requires only discovering the set of inflection slots for a single paradigm, of a single POS that must be known a priori. The presence of a word list also allows systems to anchor to a privileged form and simplifies paradigm clustering to a retrieval task.

\section{What's Next: Truly Unsupervised Paradigm Completion} \label{sec:whats-next}
\subsection{Motivation}
We introduce a version of UMPC that more strictly removes human intervention. By removing the input lexicon and evaluating more than one POS, we minimize any prior human involvement with the data and better evaluate a system's ability to generalize. This means that our only input is a raw text corpus, and it introduces two challenges. 1) We must model the entire training corpus, rather than a filtered set of words. 2) We must predict which slots to generate at test time. We design test sets to evaluate these problems, ensuring they include paradigms from at least two POS, and prompt for input forms in context, half of which are unseen in the training corpora, so systems can infer the input word POS. We refer to this version of the task as \emph{truly} unsupervised paradigm completion (\taskabbrev).

\subsection{Data and Languages} \label{sec:data}
\paragraph{Languages}
We select three development languages (English, Finnish, and Swedish) and four test languages (German, Greek, Icelandic, and Russian). We select our test languages to maximize orthographic and typological diversity, given three constraints: (1) a large number of available paradigms in UniMorph, (2) two or more POS in UniMorph, and (3) no known issues with the UniMorph data such as large numbers of missing forms. (We exclude all paradigms containing multiword forms.) We note that this yields a set of test languages that are all Indo-European, though it spans three different orthographies.

\paragraph{Raw Text Corpora}
We experiment on two corpora: the JHU Bible Corpus \cite{mccarthy-etal-2020-johns} and a child-directed corpus we create by digitizing children's books. While many studies in computational morphology focus on transcripts of child-directed speech from databases like CHILDES \cite{macwhinney2014childes}, child-directed books are part of parent's child-directed talk, and are thus an important source of language for many children  \citep{doi:10.1177/0956797615594361}. We translate the child-directed corpus into all of our languages from English using the Google Translate API following \citet{dou-neubig-2021-word}.
We tokenize with spaCy.\footnote{\url{https://spacy.io}} Details  are given in \autoref{tab:training_corpora_stats}. 

\paragraph{Test Data}
Our test data consists of words in context from two different corpora -- Wikipedia \cite{11234/1-1989} and JW300 \citep{agic-vulic-2019-jw300} --, plus their gold paradigms from UniMorph. A detailed description of the preparation of the test data can be found in \cref{app:test-set-creation}. 

\subsection{Models} \label{sec:systems}
To use existing state-of-the-art approaches and to evaluate them within the framework of \taskabbrev, we 
tackle the task with a pipeline approach, conducting 4 steps: 1) paradigm clustering, 2) slot alignment, 3) slot prediction, and 4) inflection generation. State-of-the-art models exist for Steps 1 and 4, and we propose systems for Steps 2 and 3 here, together with descriptions of those subtasks. Hyperparameters for all models are in \cref{app:hyperparams}. 

\begin{figure*}
\centering
\small
  \includegraphics[width=.75\linewidth,
  keepaspectratio]{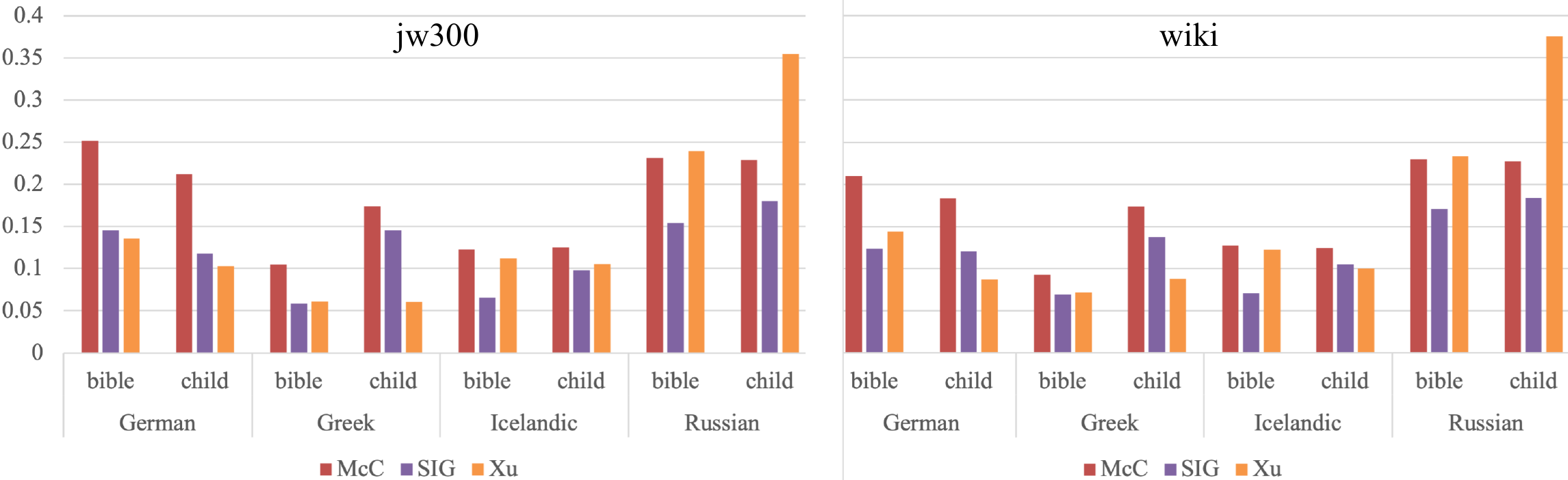}
  \caption{BMAcc for each paradigm clustering system for the POS-based slot aligner; averaged over inflectors.}
  \label{fig:f1_by_cluster}
\end{figure*}
\begin{figure*}
\centering
\small
  \includegraphics[width=.75\linewidth,
  keepaspectratio]{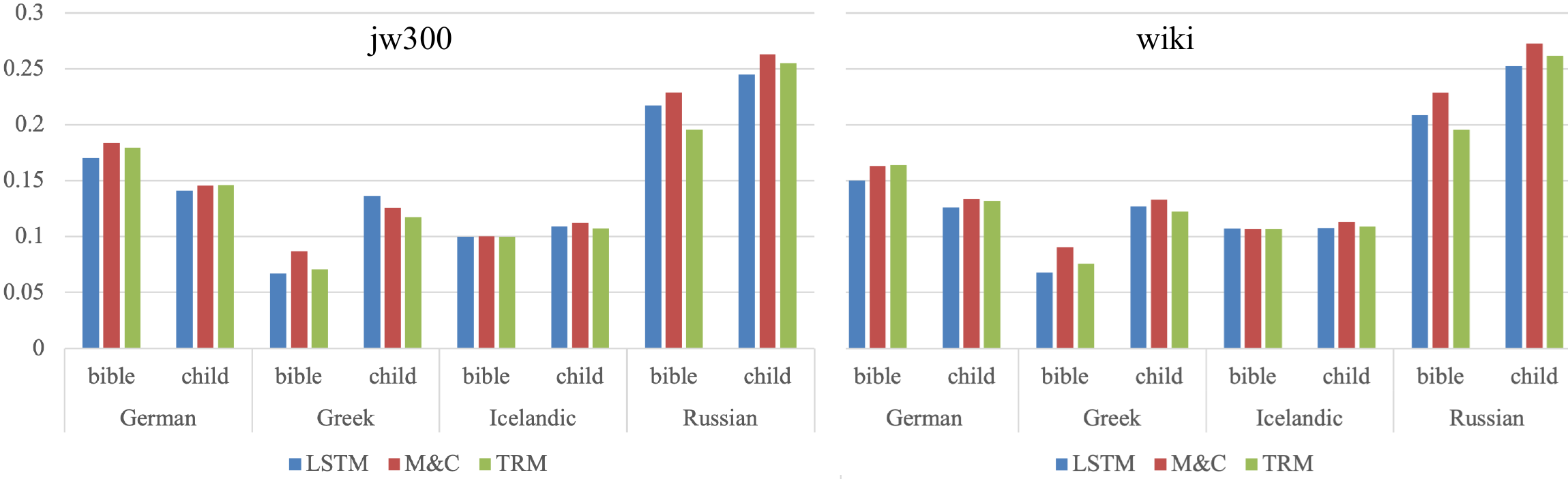}
  \caption{BMAcc for each inflector for the POS-based slot aligner; averaged over paradigm clusters.}
  \label{fig:f1_by_inflector}
\end{figure*}
\paragraph{Paradigm Clustering}
The first step for \taskabbrev{} is clustering words into paradigms. We compare 3 paradigm clustering algorithms: \citet[][\texttt{McC}]{mccurdy-etal-2021-adaptor}, \citet[][\texttt{Xu}]{xu-etal-2018-unsupervised}, and the baseline from \citet[][\texttt{SIG}]{wiemerslage-etal-2021-findings}. We modify \texttt{SIG} so it does not predict clusters which are subsets of other clusters, which improves precision. For reference, we provide those systems' paradigm clustering results in \autoref{tab:cluster_results}. In some clustering systems, each type appears in only one paradigm, which confounds our task for types that can instantiate more than one POS, and thus more than one inflectional paradigm, depending on the context.

\paragraph{Slot Alignment} \label{sec:slot-alignment}
Slot alignment is concerned with identifying which words across paradigms express the same inflectional information. 
The system we propose for the task first removes all singleton paradigm clusters from the input, as they contain no inflection pairs to learn from, and converts all remaining clusters into abstract paradigms $c_i \in C$ \cite{hulden-etal-2014-semi} by computing the longest common substring (LCS) for each cluster. For example, the LCS of the (true) paradigm of \textit{walk} is \textit{walk}, and the abstract paradigm is \textit{X0}, \textit{X0+ed}, \textit{X0+ing}, \textit{X0+s}. We filter abstract forms that appear less than $\beta = 50$ times.


Next, we assign a POS tag to each cluster. 
With a set of latent tags $Z$, we define a Bayesian model:
\begin{align}
    P(k, c_i) &= P(k) \prod_{f_j \in c_i} P(f_j \mid k) \\
    P(c_i) &= \sum_{k \in Z} P(c_i,k)
\end{align}
We then maximize the likelihood of the paradigm clusters $c_i \in C$ with an expectation maximization algorithm \cite{dempster1977maximum}. The POS assignment for each  $c_i$ is thus $\argmax_k  (P(k, c_i))$, and $|Z|$ is a hyperparameter which we set to 3.

We now have sets $C^k$. 
We assign a slot to each form in an abstract paradigm, considering one $C^k$ at a time. To this end, we compute a fastText \cite{bojanowski2017enriching} embedding for each type in the corpus and compute the embedding for an abstract form $a$ as the average fastText embedding of all types whose abstract form is $a$.
We define the similarity of two abstract forms $a$ and $a'$ as
\begin{equation}
    \operatorname{sim}(a,a') = \cos(a, a') \times (1 - J(a,a'))
    \text{,}
    \label{eq:abstract_sim}
\end{equation}
where $\cos(a, a')$ is the cosine similarity, 
 $J$ is the Jaccard similarity
\begin{equation}
    J(a, a') = \frac{|C^a \cap C^{a'}|}{|C^a \cup  C^{a'}|}\text{,}
\end{equation}
and $C^a$ is the set of abstract paradigms containing $a$. Finally, we apply agglomerative clustering over the abstract forms with \eqref{eq:abstract_sim} as our similarity metric and a distance threshold of 0.15.

\paragraph{Slot Prediction} \label{sec:slot-prediction}
Given a test form $f(\ell, \vec{t}_\gamma)$, the goal of slot prediction is to predict the source slot $\vec{t}_\gamma$ and target slots $\Gamma(\ell)$.
We treat this as a simplified POS tagging task and use a character-level Transformer seq2seq model to predict a word's POS tag and source slot. The model is trained on the results of the slot alignment step. For every word from the raw-text corpus that was assigned a slot, we sample up to 5 unique contexts. A given target word is input with its left and right neighbors; context words  that occur fewer than $\alpha = 50$ times in the training data are replaced with \textsc{oov}.
The outputs are the POS tag and the source slot generated by slot alignment. We train our model in \textsc{fairseq} \cite{ott-etal-2019-fairseq}; hyperparameters are in \cref{app:hyperparams}.



At test time, the model predicts $f(\ell, \vec{t}_\gamma)$ and the (pseudo) POS tag. Because the slot alignment step associates each POS tag with a unique set of slots, we can perform a simple lookup to find the slots that $f(\ell, \vec{t}_\gamma)$ inflects for.


\paragraph{Morphological Inflection}
To generate missing forms, we train state-of-the-art inflection models on the results of the slot alignment step and generate surface forms according to the slot prediction. 
We experiment with the following three models:
\citet[][\texttt{M\&C}]{makarov-clematide-2018-imitation}, 
\citet[][\texttt{Wu}]{wu-etal-2021-applying}, and \citet[][\texttt{K\&S}]{kann-schutze-2016-single}.



\subsection{Non-neural Baseline}
 We compare against a rule-based system (\texttt{baseline}) that heuristically predicts the same set of slots for all words, and inflects by applying edit trees to input words. A detailed description is in \cref{sec:baseline}, together with a comparison between \texttt{baseline} and our proposed POS-based system for slot alignment and slot prediction. As the POS-based system clearly outperforms \texttt{baseline}, we focus the remainder of this paper on the former.

\subsection{Results and Discussion}
 We present results from all experiments in terms of BMAcc \cite{jin-etal-2020-unsupervised}. Overall, \taskabbrev \ is difficult, though the variance in results over different components of our pipeline implies that there is a great deal of room for the community to innovate. We see the lowest scores for our Greek and Icelandic corpora. These have far fewer tokens than German and Russian, plus higher type--token ratios, which likely makes the task more challenging.

\paragraph{Impact of the Clustering System} \Cref{fig:f1_by_cluster} shows that the choice of paradigm clustering strategy strongly affects our pipeline's downstream performance. \texttt{McC}, the best performing clustering system on the paradigm clustering task, frequently outperforms the other two strategies. The exception to this is Russian, where \texttt{Xu} gives the best results---by a large margin when learning from the child-directed training corpora. 

\paragraph{Impact of the Inflection System} From \autoref{fig:f1_by_inflector} it is obvious that the choice of inflection model does \textit{not} have a large effect on downstream results. All three systems we compare are known to be extremely competitive on the supervised inflection task, so it is reasonable to assume that they fit the generated training data relatively similarly. Future work can assess how inflection generation can best account for the noisy nature of the data in this task, akin to \citet{michel-neubig-2018-mtnt}.

\paragraph{Impact of the Corpus} The consilience of our results suggests that the child-directed corpus leads to slightly better downstream performance, except in German. Notably, the German Bible contains far more tokens and far fewer types than the corresponding child-directed corpus (\autoref{tab:training_corpora_stats}), which may significantly simplify the learning task.

\section{Conclusion}
Thanks to strong systems for inflection, segmentation, and paradigm completion, computational morphology is ripe to contribute to the large number of the world’s languages with very few digital resources. We explore this through the novel task \taskabbrev{}---which presents several challenges. We believe that truly unsupervised morphology is an important direction, and it can have a large impact on language technology for thousands of languages.
With the goal of preserving endangered languages, we note that more than half the world's languages have no writing system \cite{harmon1995status}. A frontier for this task would process speech as a strategy for language documentation in unwritten languages.

\section*{Acknowledgments} We thank David Yarowsky, members of NALA, and the anonymous reviewers for helpful feedback.
\bibliographystyle{acl_natbib}
\bibliography{anthology,acl2021,custom}

\begin{thebibliography}{155}
\expandafter\ifx\csname natexlab\endcsname\relax\def\natexlab#1{#1}\fi

\bibitem[{Ackerman et~al.(2009)Ackerman, Blevins, and
  Malouf}]{ackerman2009parts}
Farrell Ackerman, James~P Blevins, and Robert Malouf. 2009.
\newblock Parts and wholes: Implicative patterns in inflectional paradigms.
\newblock \emph{Analogy in grammar: Form and acquisition}, pages 54--82.

\bibitem[{{\'A}cs(2018)}]{acs-2018-bme}
Judit {\'A}cs. 2018.
\newblock \href {https://doi.org/10.18653/v1/K18-3016} {{BME}-{HAS} system for
  {C}o{NLL}{--}{SIGMORPHON} 2018 shared task: Universal morphological
  reinflection}.
\newblock In \emph{Proceedings of the {C}o{NLL}{--}{SIGMORPHON} 2018 Shared
  Task: Universal Morphological Reinflection}, pages 121--126, Brussels.
  Association for Computational Linguistics.

\bibitem[{Agi{\'c} and Vuli{\'c}(2019)}]{agic-vulic-2019-jw300}
{\v{Z}}eljko Agi{\'c} and Ivan Vuli{\'c}. 2019.
\newblock \href {https://doi.org/10.18653/v1/P19-1310} {{JW}300: A
  wide-coverage parallel corpus for low-resource languages}.
\newblock In \emph{Proceedings of the 57th Annual Meeting of the Association
  for Computational Linguistics}, pages 3204--3210, Florence, Italy.
  Association for Computational Linguistics.

\bibitem[{Aharoni and Goldberg(2017)}]{aharoni-goldberg-2017-morphological}
Roee Aharoni and Yoav Goldberg. 2017.
\newblock \href {https://doi.org/10.18653/v1/P17-1183} {Morphological
  inflection generation with hard monotonic attention}.
\newblock In \emph{Proceedings of the 55th Annual Meeting of the Association
  for Computational Linguistics (Volume 1: Long Papers)}, pages 2004--2015,
  Vancouver, Canada. Association for Computational Linguistics.

\bibitem[{Anastasopoulos and Neubig(2019)}]{anastasopoulos-neubig-2019-pushing}
Antonios Anastasopoulos and Graham Neubig. 2019.
\newblock \href {https://doi.org/10.18653/v1/D19-1091} {Pushing the limits of
  low-resource morphological inflection}.
\newblock In \emph{Proceedings of the 2019 Conference on Empirical Methods in
  Natural Language Processing and the 9th International Joint Conference on
  Natural Language Processing (EMNLP-IJCNLP)}, pages 984--996, Hong Kong,
  China. Association for Computational Linguistics.

\bibitem[{Arppe et~al.(2021)Arppe, Good, Harrigan, Hulden, Lachler, Moeller,
  Palmer, Silfverberg, and Schwartz}]{computel-2021-use}
Antti Arppe, Jeff Good, Atticus Harrigan, Mans Hulden, Jordan Lachler, Sarah
  Moeller, Alexis Palmer, Miikka Silfverberg, and Lane Schwartz, editors. 2021.
\newblock \href {https://aclanthology.org/2021.computel-1.0} {\emph{Proceedings
  of the 4th Workshop on the Use of Computational Methods in the Study of
  Endangered Languages Volume 1 (Papers)}}. Association for Computational
  Linguistics, Online.

\bibitem[{Attardi(2015)}]{Wikiextractor2015}
Giusepppe Attardi. 2015.
\newblock Wikiextractor.
\newblock \url{https://github.com/attardi/wikiextractor}.

\bibitem[{Austin and Sallabank(2011)}]{austin_sallabank_2011}
P.~Austin and J.~Sallabank, editors. 2011.
\newblock \emph{The Cambridge Handbook of Endangered Languages}.
\newblock Cambridge Handbooks in Language and Linguistics. Cambridge University
  Press.

\bibitem[{Baayen et~al.(1996)Baayen, Piepenbrock, and
  Gulikers}]{baayen1996celex}
R.~Harald Baayen, Richard Piepenbrock, and Leon Gulikers. 1996.
\newblock The {CELEX} lexical database ({CD-ROM}).

\bibitem[{Bahdanau et~al.(2015)Bahdanau, Cho, and Bengio}]{bahdanau-et-al}
Dzmitry Bahdanau, Kyunghyun Cho, and Yoshua Bengio. 2015.
\newblock Neural machine translation by jointly learning to align and
  translate.
\newblock \emph{CoRR}, abs/1409.0473.

\bibitem[{Bergmanis and Goldwater(2017)}]{bergmanis2017segmentation}
Toms Bergmanis and Sharon Goldwater. 2017.
\newblock \href {https://aclanthology.org/E17-1032} {From segmentation to
  analyses: a probabilistic model for unsupervised morphology induction}.
\newblock In \emph{Proceedings of the 15th Conference of the {E}uropean Chapter
  of the Association for Computational Linguistics: Volume 1, Long Papers},
  pages 337--346, Valencia, Spain. Association for Computational Linguistics.

\bibitem[{Bergmanis and Goldwater(2018)}]{bergmanis-goldwater-2018-context}
Toms Bergmanis and Sharon Goldwater. 2018.
\newblock \href {https://doi.org/10.18653/v1/N18-1126} {Context sensitive
  neural lemmatization with {L}ematus}.
\newblock In \emph{Proceedings of the 2018 Conference of the North {A}merican
  Chapter of the Association for Computational Linguistics: Human Language
  Technologies, Volume 1 (Long Papers)}, pages 1391--1400, New Orleans,
  Louisiana. Association for Computational Linguistics.

\bibitem[{Bergmanis and Goldwater(2019)}]{bergmanis-goldwater-2019-data}
Toms Bergmanis and Sharon Goldwater. 2019.
\newblock \href {https://doi.org/10.18653/v1/N19-1418} {Data augmentation for
  context-sensitive neural lemmatization using inflection tables and raw text}.
\newblock In \emph{Proceedings of the 2019 Conference of the North {A}merican
  Chapter of the Association for Computational Linguistics: Human Language
  Technologies, Volume 1 (Long and Short Papers)}, pages 4119--4128,
  Minneapolis, Minnesota. Association for Computational Linguistics.

\bibitem[{Bergmanis et~al.(2017)Bergmanis, Kann, Sch{\"u}tze, and
  Goldwater}]{bergmanis-etal-2017-training}
Toms Bergmanis, Katharina Kann, Hinrich Sch{\"u}tze, and Sharon Goldwater.
  2017.
\newblock \href {https://doi.org/10.18653/v1/K17-2002} {Training data
  augmentation for low-resource morphological inflection}.
\newblock In \emph{Proceedings of the {C}o{NLL} {SIGMORPHON} 2017 Shared Task:
  Universal Morphological Reinflection}, pages 31--39, Vancouver. Association
  for Computational Linguistics.

\bibitem[{Blevins and Zettlemoyer(2019)}]{blevins-zettlemoyer-2019-better}
Terra Blevins and Luke Zettlemoyer. 2019.
\newblock \href {https://doi.org/10.18653/v1/P19-1156} {Better character
  language modeling through morphology}.
\newblock In \emph{Proceedings of the 57th Annual Meeting of the Association
  for Computational Linguistics}, pages 1606--1613, Florence, Italy.
  Association for Computational Linguistics.

\bibitem[{Bojanowski et~al.(2017)Bojanowski, Grave, Joulin, and
  Mikolov}]{bojanowski2017enriching}
Piotr Bojanowski, Edouard Grave, Armand Joulin, and Tomas Mikolov. 2017.
\newblock Enriching word vectors with subword information.
\newblock \emph{Transactions of the Association for Computational Linguistics},
  5:135--146.

\bibitem[{Buys and Botha(2016)}]{buys-botha-2016-cross}
Jan Buys and Jan~A. Botha. 2016.
\newblock \href {https://doi.org/10.18653/v1/P16-1184} {Cross-lingual
  morphological tagging for low-resource languages}.
\newblock In \emph{Proceedings of the 54th Annual Meeting of the Association
  for Computational Linguistics (Volume 1: Long Papers)}, pages 1954--1964,
  Berlin, Germany. Association for Computational Linguistics.

\bibitem[{Calzolari et~al.(2020)Calzolari, B{\'e}chet, Blache, Choukri, Cieri,
  Declerck, Goggi, Isahara, Maegaard, Mariani, Mazo, Moreno, Odijk, and
  Piperidis}]{lrec-2020-language}
Nicoletta Calzolari, Fr{\'e}d{\'e}ric B{\'e}chet, Philippe Blache, Khalid
  Choukri, Christopher Cieri, Thierry Declerck, Sara Goggi, Hitoshi Isahara,
  Bente Maegaard, Joseph Mariani, et~al., editors. 2020.
\newblock \href {https://www.aclweb.org/anthology/2020.lrec-1.0}
  {\emph{Proceedings of the 12th Language Resources and Evaluation
  Conference}}. European Language Resources Association, Marseille, France.

\bibitem[{Chan(2006)}]{chan-2006-learning}
Erwin Chan. 2006.
\newblock \href {https://www.aclweb.org/anthology/W06-3209} {Learning
  probabilistic paradigms for morphology in a latent class model}.
\newblock In \emph{Proceedings of the Eighth Meeting of the {ACL} Special
  Interest Group on Computational Phonology and Morphology at {HLT}-{NAACL}
  2006}, pages 69--78, New York City, USA. Association for Computational
  Linguistics.

\bibitem[{Chrupa{\l}a(2008)}]{chrupala2008towards}
Grzegorz Chrupa{\l}a. 2008.
\newblock \emph{Towards a machine-learning architecture for lexical functional
  grammar parsing}.
\newblock Ph.D. thesis, Dublin City University.

\bibitem[{Church(2020)}]{DBLP:journals/nle/Church20a}
Kenneth~Ward Church. 2020.
\newblock \href {https://doi.org/10.1017/S1351324920000145} {Emerging trends:
  Subwords, seriously?}
\newblock \emph{Nat. Lang. Eng.}, 26(3):375--382.

\bibitem[{Clark(2001)}]{Clark2001LearningMW}
Alexander Clark. 2001.
\newblock Learning morphology with pair hidden {M}arkov models.
\newblock In \emph{ACL (Companion Volume)}.

\bibitem[{Cotterell and Heigold(2017)}]{cotterell-heigold-2017-cross}
Ryan Cotterell and Georg Heigold. 2017.
\newblock \href {https://doi.org/10.18653/v1/D17-1078} {Cross-lingual
  character-level neural morphological tagging}.
\newblock In \emph{Proceedings of the 2017 Conference on Empirical Methods in
  Natural Language Processing}, pages 748--759, Copenhagen, Denmark.
  Association for Computational Linguistics.

\bibitem[{Cotterell et~al.(2018)Cotterell, Kirov, Sylak-Glassman, Walther,
  Vylomova, McCarthy, Kann, Mielke, Nicolai, Silfverberg, Yarowsky, Eisner, and
  Hulden}]{cotterell-etal-2018-conll}
Ryan Cotterell, Christo Kirov, John Sylak-Glassman, G{\'e}raldine Walther,
  Ekaterina Vylomova, Arya~D. McCarthy, Katharina Kann, Sabrina~J. Mielke,
  Garrett Nicolai, Miikka Silfverberg, et~al. 2018.
\newblock \href {https://doi.org/10.18653/v1/K18-3001} {The
  {C}o{NLL}{--}{SIGMORPHON} 2018 shared task: Universal morphological
  reinflection}.
\newblock In \emph{Proceedings of the {C}o{NLL}{--}{SIGMORPHON} 2018 Shared
  Task: Universal Morphological Reinflection}, pages 1--27, Brussels.
  Association for Computational Linguistics.

\bibitem[{Cotterell et~al.(2017{\natexlab{a}})Cotterell, Kirov, Sylak-Glassman,
  Walther, Vylomova, Xia, Faruqui, K{\"u}bler, Yarowsky, Eisner, and
  Hulden}]{cotterell-etal-2017-conll}
Ryan Cotterell, Christo Kirov, John Sylak-Glassman, G{\'e}raldine Walther,
  Ekaterina Vylomova, Patrick Xia, Manaal Faruqui, Sandra K{\"u}bler, David
  Yarowsky, Jason Eisner, and Mans Hulden. 2017{\natexlab{a}}.
\newblock \href {https://doi.org/10.18653/v1/K17-2001} {{C}o{NLL}-{SIGMORPHON}
  2017 shared task: Universal morphological reinflection in 52 languages}.
\newblock In \emph{Proceedings of the {C}o{NLL} {SIGMORPHON} 2017 Shared Task:
  Universal Morphological Reinflection}, pages 1--30, Vancouver. Association
  for Computational Linguistics.

\bibitem[{Cotterell et~al.(2016{\natexlab{a}})Cotterell, Kirov, Sylak-Glassman,
  Yarowsky, Eisner, and Hulden}]{cotterell-etal-2016-sigmorphon}
Ryan Cotterell, Christo Kirov, John Sylak-Glassman, David Yarowsky, Jason
  Eisner, and Mans Hulden. 2016{\natexlab{a}}.
\newblock \href {https://doi.org/10.18653/v1/W16-2002} {The {SIGMORPHON} 2016
  shared {T}ask{---}{M}orphological reinflection}.
\newblock In \emph{Proceedings of the 14th {SIGMORPHON} Workshop on
  Computational Research in Phonetics, Phonology, and Morphology}, pages
  10--22, Berlin, Germany. Association for Computational Linguistics.

\bibitem[{Cotterell et~al.(2016{\natexlab{b}})Cotterell, Kirov, Sylak-Glassman,
  Yarowsky, Eisner, and Hulden}]{cotterell2016sigmorphon}
Ryan Cotterell, Christo Kirov, John Sylak-Glassman, David Yarowsky, Jason
  Eisner, and Mans Hulden. 2016{\natexlab{b}}.
\newblock The {SIGMORPHON} 2016 shared task—morphological reinflection.
\newblock In \emph{Proceedings of the 14th SIGMORPHON Workshop on Computational
  Research in Phonetics, Phonology, and Morphology}, pages 10--22.

\bibitem[{Cotterell et~al.(2015)Cotterell, M{\"u}ller, Fraser, and
  Sch{\"u}tze}]{cotterell-etal-2015-labeled}
Ryan Cotterell, Thomas M{\"u}ller, Alexander Fraser, and Hinrich Sch{\"u}tze.
  2015.
\newblock \href {https://doi.org/10.18653/v1/K15-1017} {Labeled morphological
  segmentation with semi-{M}arkov models}.
\newblock In \emph{Proceedings of the Nineteenth Conference on Computational
  Natural Language Learning}, pages 164--174, Beijing, China. Association for
  Computational Linguistics.

\bibitem[{Cotterell et~al.(2017{\natexlab{b}})Cotterell, Sylak-Glassman, and
  Kirov}]{cotterell-etal-2017-neural}
Ryan Cotterell, John Sylak-Glassman, and Christo Kirov. 2017{\natexlab{b}}.
\newblock \href {https://www.aclweb.org/anthology/E17-2120} {Neural graphical
  models over strings for principal parts morphological paradigm completion}.
\newblock In \emph{Proceedings of the 15th Conference of the {E}uropean Chapter
  of the Association for Computational Linguistics: Volume 2, Short Papers},
  pages 759--765, Valencia, Spain. Association for Computational Linguistics.

\bibitem[{Cotterell et~al.(2016{\natexlab{c}})Cotterell, Vieira, and
  Sch{\"u}tze}]{cotterell-etal-2016-joint}
Ryan Cotterell, Tim Vieira, and Hinrich Sch{\"u}tze. 2016{\natexlab{c}}.
\newblock \href {https://doi.org/10.18653/v1/N16-1080} {A joint model of
  orthography and morphological segmentation}.
\newblock In \emph{Proceedings of the 2016 Conference of the North {A}merican
  Chapter of the Association for Computational Linguistics: Human Language
  Technologies}, pages 664--669, San Diego, California. Association for
  Computational Linguistics.

\bibitem[{Creutz and Lagus(2002)}]{creutz-lagus-2002-unsupervised}
Mathias Creutz and Krista Lagus. 2002.
\newblock \href {https://doi.org/10.3115/1118647.1118650} {Unsupervised
  discovery of morphemes}.
\newblock In \emph{Proceedings of the {ACL}-02 Workshop on Morphological and
  Phonological Learning}, pages 21--30. Association for Computational
  Linguistics.

\bibitem[{Creutz and Lagus(2005)}]{creutz2005inducing}
Mathias Creutz and Krista Lagus. 2005.
\newblock Inducing the morphological lexicon of a natural language from
  unannotated text.
\newblock In \emph{Proceedings of the International and Interdisciplinary
  Conference on Adaptive Knowledge Representation and Reasoning (AKRR’05)},
  106-113, pages 51--59.

\bibitem[{Dempster et~al.(1977)Dempster, Laird, and
  Rubin}]{dempster1977maximum}
Arthur~P. Dempster, Nan~M. Laird, and Donald~B. Rubin. 1977.
\newblock Maximum likelihood from incomplete data via the {EM} algorithm.
\newblock \emph{Journal of the Royal Statistical Society: Series B
  (Methodological)}, 39(1):1--22.

\bibitem[{Devlin et~al.(2019)Devlin, Chang, Lee, and
  Toutanova}]{devlin-etal-2019-bert}
Jacob Devlin, Ming-Wei Chang, Kenton Lee, and Kristina Toutanova. 2019.
\newblock \href {https://doi.org/10.18653/v1/N19-1423} {{BERT}: Pre-training of
  deep bidirectional transformers for language understanding}.
\newblock In \emph{Proceedings of the 2019 Conference of the North {A}merican
  Chapter of the Association for Computational Linguistics: Human Language
  Technologies, Volume 1 (Long and Short Papers)}, pages 4171--4186,
  Minneapolis, Minnesota. Association for Computational Linguistics.

\bibitem[{Dou and Neubig(2021)}]{dou-neubig-2021-word}
Zi-Yi Dou and Graham Neubig. 2021.
\newblock \href {https://doi.org/10.18653/v1/2021.eacl-main.181} {Word
  alignment by fine-tuning embeddings on parallel corpora}.
\newblock In \emph{Proceedings of the 16th Conference of the European Chapter
  of the Association for Computational Linguistics: Main Volume}, pages
  2112--2128, Online. Association for Computational Linguistics.

\bibitem[{Durrett and DeNero(2013)}]{durrett-denero-2013-supervised}
Greg Durrett and John DeNero. 2013.
\newblock \href {https://www.aclweb.org/anthology/N13-1138} {Supervised
  learning of complete morphological paradigms}.
\newblock In \emph{Proceedings of the 2013 Conference of the North {A}merican
  Chapter of the Association for Computational Linguistics: Human Language
  Technologies}, pages 1185--1195, Atlanta, Georgia. Association for
  Computational Linguistics.

\bibitem[{Dyer et~al.(2008)Dyer, Muresan, and
  Resnik}]{dyer-etal-2008-generalizing}
Christopher Dyer, Smaranda Muresan, and Philip Resnik. 2008.
\newblock \href {https://www.aclweb.org/anthology/P08-1115} {Generalizing word
  lattice translation}.
\newblock In \emph{Proceedings of ACL-08: HLT}, pages 1012--1020, Columbus,
  Ohio. Association for Computational Linguistics.

\bibitem[{Elsner et~al.(2019)Elsner, Sims, Erdmann, Hernandez, Jaffe, Jin,
  Johnson, Karim, King, Nunes et~al.}]{elsner2019modeling}
Micha Elsner, Andrea~D Sims, Alexander Erdmann, Antonio Hernandez, Evan Jaffe,
  Lifeng Jin, Martha~Booker Johnson, Shuan Karim, David~L King, Luana~Lamberti
  Nunes, et~al. 2019.
\newblock Modeling morphological learning, typology, and change: What can the
  neural sequence-to-sequence framework contribute?
\newblock \emph{Journal of Language Modelling}, 7(1):53--98.

\bibitem[{Erdmann et~al.(2020)Erdmann, Elsner, Wu, Cotterell, and
  Habash}]{erdmann-etal-2020-paradigm}
Alexander Erdmann, Micha Elsner, Shijie Wu, Ryan Cotterell, and Nizar Habash.
  2020.
\newblock \href {https://doi.org/10.18653/v1/2020.acl-main.695} {The paradigm
  discovery problem}.
\newblock In \emph{Proceedings of the 58th Annual Meeting of the Association
  for Computational Linguistics}, pages 7778--7790, Online. Association for
  Computational Linguistics.

\bibitem[{Eskander et~al.(2020)Eskander, Callejas, Nichols, Klavans, and
  Muresan}]{eskander-etal-2020-morphagram}
Ramy Eskander, Francesca Callejas, Elizabeth Nichols, Judith Klavans, and
  Smaranda Muresan. 2020.
\newblock \href {https://www.aclweb.org/anthology/2020.lrec-1.879}
  {{M}orph{AG}ram, evaluation and framework for unsupervised morphological
  segmentation}.
\newblock In \emph{Proceedings of the 12th Language Resources and Evaluation
  Conference}, pages 7112--7122, Marseille, France. European Language Resources
  Association.

\bibitem[{Eskander et~al.(2019)Eskander, Klavans, and
  Muresan}]{eskander-etal-2019-unsupervised}
Ramy Eskander, Judith Klavans, and Smaranda Muresan. 2019.
\newblock \href {https://doi.org/10.18653/v1/W19-4222} {Unsupervised
  morphological segmentation for low-resource polysynthetic languages}.
\newblock In \emph{Proceedings of the 16th Workshop on Computational Research
  in Phonetics, Phonology, and Morphology}, pages 189--195, Florence, Italy.
  Association for Computational Linguistics.

\bibitem[{Faruqui et~al.(2016)Faruqui, Tsvetkov, Neubig, and
  Dyer}]{faruqui-etal-2016-morphological}
Manaal Faruqui, Yulia Tsvetkov, Graham Neubig, and Chris Dyer. 2016.
\newblock \href {https://doi.org/10.18653/v1/N16-1077} {Morphological
  inflection generation using character sequence to sequence learning}.
\newblock In \emph{Proceedings of the 2016 Conference of the North {A}merican
  Chapter of the Association for Computational Linguistics: Human Language
  Technologies}, pages 634--643, San Diego, California. Association for
  Computational Linguistics.

\bibitem[{Finkel and Stump(2007)}]{finkel2007principal}
Raphael Finkel and Gregory Stump. 2007.
\newblock Principal parts and morphological typology.
\newblock \emph{Morphology}, 17(1):39--75.

\bibitem[{Ginter et~al.(2017)Ginter, Haji{\v c}, Luotolahti, Straka, and
  Zeman}]{11234/1-1989}
Filip Ginter, Jan Haji{\v c}, Juhani Luotolahti, Milan Straka, and Daniel
  Zeman. 2017.
\newblock \href {http://hdl.handle.net/11234/1-1989} {{CoNLL} 2017 shared task
  - automatically annotated raw texts and word embeddings}.
\newblock {LINDAT}/{CLARIAH}-{CZ} digital library at the Institute of Formal
  and Applied Linguistics ({{\'U}FAL}), Faculty of Mathematics and Physics,
  Charles University.

\bibitem[{Goldman and Tsarfaty(2021)}]{goldman-tsarfaty-2021-well}
Omer Goldman and Reut Tsarfaty. 2021.
\newblock \href {https://aclanthology.org/2021.mrl-1.23} {Well-defined
  morphology is sentence-level morphology}.
\newblock In \emph{Proceedings of the 1st Workshop on Multilingual
  Representation Learning}, pages 248--250, Punta Cana, Dominican Republic.
  Association for Computational Linguistics.

\bibitem[{Goldsmith(2001)}]{goldsmith-2001-unsupervised}
John Goldsmith. 2001.
\newblock \href {https://doi.org/10.1162/089120101750300490} {Unsupervised
  learning of the morphology of a natural language}.
\newblock \emph{Computational Linguistics}, 27(2):153--198.

\bibitem[{Goldsmith(2010)}]{goldsmith2010segmentation}
John~A Goldsmith. 2010.
\newblock Segmentation and morphology.
\newblock \emph{The handbook of computational linguistics and natural language
  processing}, 57:364.

\bibitem[{Gorman et~al.(2019)Gorman, McCarthy, Cotterell, Vylomova,
  Silfverberg, and Markowska}]{gorman-etal-2019-weird}
Kyle Gorman, Arya~D. McCarthy, Ryan Cotterell, Ekaterina Vylomova, Miikka
  Silfverberg, and Magdalena Markowska. 2019.
\newblock \href {https://doi.org/10.18653/v1/K19-1014} {Weird inflects but
  {OK}: Making sense of morphological generation errors}.
\newblock In \emph{Proceedings of the 23rd Conference on Computational Natural
  Language Learning (CoNLL)}, pages 140--151, Hong Kong, China. Association for
  Computational Linguistics.

\bibitem[{Gr{\"o}nroos et~al.(2019)Gr{\"o}nroos, Virpioja, and
  Kurimo}]{gronroos-etal-2019-north}
Stig-Arne Gr{\"o}nroos, S{\'a}mi Virpioja, and Mikko Kurimo. 2019.
\newblock \href {https://doi.org/10.18653/v1/W19-0302} {{N}orth {S}{\'a}mi
  morphological segmentation with low-resource semi-supervised sequence
  labeling}.
\newblock In \emph{Proceedings of the Fifth International Workshop on
  Computational Linguistics for Uralic Languages}, pages 15--26, Tartu,
  Estonia. Association for Computational Linguistics.

\bibitem[{Haji{\v{c}} and Zeman(2017)}]{conll-2017-conll-2017}
Jan Haji{\v{c}} and Dan Zeman, editors. 2017.
\newblock \href {https://doi.org/10.18653/v1/K17-3} {\emph{Proceedings of the
  {C}o{NLL} 2017 Shared Task: Multilingual Parsing from Raw Text to Universal
  Dependencies}}. Association for Computational Linguistics, Vancouver, Canada.

\bibitem[{Harmon(1995)}]{harmon1995status}
David Harmon. 1995.
\newblock The status of the world's languages as reported in ``{E}thnologue".
\newblock \emph{Southwest Journal of Linguistics}, 14:1--28.

\bibitem[{Harris(1970)}]{harris1970morpheme}
Zellig~S Harris. 1970.
\newblock Morpheme boundaries within words: Report on a computer test.
\newblock In \emph{Papers in Structural and Transformational Linguistics},
  pages 68--77. Springer.

\bibitem[{Heigold et~al.(2016)Heigold, Neumann, and van
  Genabith}]{heigold2016neural}
Georg Heigold, Guenter Neumann, and Josef van Genabith. 2016.
\newblock Neural morphological tagging from characters for morphologically rich
  languages.
\newblock \emph{arXiv preprint arXiv:1606.06640}.

\bibitem[{Heigold et~al.(2017)Heigold, Neumann, and van
  Genabith}]{heigold-etal-2017-extensive}
Georg Heigold, Guenter Neumann, and Josef van Genabith. 2017.
\newblock \href {https://www.aclweb.org/anthology/E17-1048} {An extensive
  empirical evaluation of character-based morphological tagging for 14
  languages}.
\newblock In \emph{Proceedings of the 15th Conference of the {E}uropean Chapter
  of the Association for Computational Linguistics: Volume 1, Long Papers},
  pages 505--513, Valencia, Spain. Association for Computational Linguistics.

\bibitem[{Hochreiter and Schmidhuber(1997)}]{hochreiter1997long}
Sepp Hochreiter and J{\"u}rgen Schmidhuber. 1997.
\newblock Long short-term memory.
\newblock \emph{Neural computation}, 9(8):1735--1780.

\bibitem[{Hofmann et~al.(2021)Hofmann, Pierrehumbert, and
  Sch{\"u}tze}]{hofmann-etal-2021-superbizarre}
Valentin Hofmann, Janet Pierrehumbert, and Hinrich Sch{\"u}tze. 2021.
\newblock \href {https://doi.org/10.18653/v1/2021.acl-long.279} {Superbizarre
  is not superb: Derivational morphology improves {BERT}{'}s interpretation of
  complex words}.
\newblock In \emph{Proceedings of the 59th Annual Meeting of the Association
  for Computational Linguistics and the 11th International Joint Conference on
  Natural Language Processing (Volume 1: Long Papers)}, pages 3594--3608,
  Online. Association for Computational Linguistics.

\bibitem[{Hohensee and Bender(2012)}]{hohensee-bender-2012-getting}
Matt Hohensee and Emily~M. Bender. 2012.
\newblock \href {https://www.aclweb.org/anthology/N12-1032} {Getting more from
  morphology in multilingual dependency parsing}.
\newblock In \emph{Proceedings of the 2012 Conference of the North {A}merican
  Chapter of the Association for Computational Linguistics: Human Language
  Technologies}, pages 315--326, Montr{\'e}al, Canada. Association for
  Computational Linguistics.

\bibitem[{Hulden et~al.(2014)Hulden, Forsberg, and
  Ahlberg}]{hulden-etal-2014-semi}
Mans Hulden, Markus Forsberg, and Malin Ahlberg. 2014.
\newblock \href {https://doi.org/10.3115/v1/E14-1060} {Semi-supervised learning
  of morphological paradigms and lexicons}.
\newblock In \emph{Proceedings of the 14th Conference of the {E}uropean Chapter
  of the Association for Computational Linguistics}, pages 569--578,
  Gothenburg, Sweden. Association for Computational Linguistics.

\bibitem[{Jin et~al.(2020)Jin, Cai, Peng, Xia, McCarthy, and
  Kann}]{jin-etal-2020-unsupervised}
Huiming Jin, Liwei Cai, Yihui Peng, Chen Xia, Arya McCarthy, and Katharina
  Kann. 2020.
\newblock \href {https://doi.org/10.18653/v1/2020.acl-main.598} {Unsupervised
  morphological paradigm completion}.
\newblock In \emph{Proceedings of the 58th Annual Meeting of the Association
  for Computational Linguistics}, pages 6696--6707, Online. Association for
  Computational Linguistics.

\bibitem[{Johnson et~al.(2007)Johnson, Griffiths, Goldwater
  et~al.}]{johnson2007adaptor}
Mark Johnson, Thomas~L Griffiths, Sharon Goldwater, et~al. 2007.
\newblock Adaptor grammars: A framework for specifying compositional
  nonparametric bayesian models.
\newblock \emph{Advances in neural information processing systems}, 19:641.

\bibitem[{Kann et~al.(2016)Kann, Cotterell, and
  Sch{\"u}tze}]{kann-etal-2016-neural}
Katharina Kann, Ryan Cotterell, and Hinrich Sch{\"u}tze. 2016.
\newblock \href {https://doi.org/10.18653/v1/D16-1097} {Neural morphological
  analysis: Encoding-decoding canonical segments}.
\newblock In \emph{Proceedings of the 2016 Conference on Empirical Methods in
  Natural Language Processing}, pages 961--967, Austin, Texas. Association for
  Computational Linguistics.

\bibitem[{Kann et~al.(2017{\natexlab{a}})Kann, Cotterell, and
  Sch{\"u}tze}]{kann-etal-2017-neural}
Katharina Kann, Ryan Cotterell, and Hinrich Sch{\"u}tze. 2017{\natexlab{a}}.
\newblock \href {https://www.aclweb.org/anthology/E17-1049} {Neural
  multi-source morphological reinflection}.
\newblock In \emph{Proceedings of the 15th Conference of the {E}uropean Chapter
  of the Association for Computational Linguistics: Volume 1, Long Papers},
  pages 514--524, Valencia, Spain. Association for Computational Linguistics.

\bibitem[{Kann et~al.(2017{\natexlab{b}})Kann, Cotterell, and
  Sch{\"u}tze}]{kann-etal-2017-one}
Katharina Kann, Ryan Cotterell, and Hinrich Sch{\"u}tze. 2017{\natexlab{b}}.
\newblock \href {https://doi.org/10.18653/v1/P17-1182} {One-shot neural
  cross-lingual transfer for paradigm completion}.
\newblock In \emph{Proceedings of the 55th Annual Meeting of the Association
  for Computational Linguistics (Volume 1: Long Papers)}, pages 1993--2003,
  Vancouver, Canada. Association for Computational Linguistics.

\bibitem[{Kann et~al.(2018{\natexlab{a}})Kann, Lauly, and
  Cho}]{kann-etal-2018-nyu}
Katharina Kann, Stanislas Lauly, and Kyunghyun Cho. 2018{\natexlab{a}}.
\newblock \href {https://doi.org/10.18653/v1/K18-3006} {The {NYU} system for
  the {C}o{NLL}{--}{SIGMORPHON} 2018 shared task on universal morphological
  reinflection}.
\newblock In \emph{Proceedings of the {C}o{NLL}{--}{SIGMORPHON} 2018 Shared
  Task: Universal Morphological Reinflection}, pages 58--63, Brussels.
  Association for Computational Linguistics.

\bibitem[{Kann et~al.(2018{\natexlab{b}})Kann, Mager~Hois, Meza-Ruiz, and
  Sch{\"u}tze}]{kann-etal-2018-fortification}
Katharina Kann, Jesus~Manuel Mager~Hois, Ivan~Vladimir Meza-Ruiz, and Hinrich
  Sch{\"u}tze. 2018{\natexlab{b}}.
\newblock \href {https://doi.org/10.18653/v1/N18-1005} {Fortification of neural
  morphological segmentation models for polysynthetic minimal-resource
  languages}.
\newblock In \emph{Proceedings of the 2018 Conference of the North {A}merican
  Chapter of the Association for Computational Linguistics: Human Language
  Technologies, Volume 1 (Long Papers)}, pages 47--57, New Orleans, Louisiana.
  Association for Computational Linguistics.

\bibitem[{Kann et~al.(2020)Kann, McCarthy, Nicolai, and
  Hulden}]{kann-etal-2020-sigmorphon}
Katharina Kann, Arya~D. McCarthy, Garrett Nicolai, and Mans Hulden. 2020.
\newblock \href {https://doi.org/10.18653/v1/2020.sigmorphon-1.3} {The
  {SIGMORPHON} 2020 shared task on unsupervised morphological paradigm
  completion}.
\newblock In \emph{Proceedings of the 17th SIGMORPHON Workshop on Computational
  Research in Phonetics, Phonology, and Morphology}, pages 51--62, Online.
  Association for Computational Linguistics.

\bibitem[{Kann and Sch{\"u}tze(2016{\natexlab{a}})}]{kann-schutze-2016-med}
Katharina Kann and Hinrich Sch{\"u}tze. 2016{\natexlab{a}}.
\newblock \href {https://doi.org/10.18653/v1/W16-2010} {{MED}: The {LMU} system
  for the {SIGMORPHON} 2016 shared task on morphological reinflection}.
\newblock In \emph{Proceedings of the 14th {SIGMORPHON} Workshop on
  Computational Research in Phonetics, Phonology, and Morphology}, pages
  62--70, Berlin, Germany. Association for Computational Linguistics.

\bibitem[{Kann and Sch{\"u}tze(2016{\natexlab{b}})}]{kann-schutze-2016-single}
Katharina Kann and Hinrich Sch{\"u}tze. 2016{\natexlab{b}}.
\newblock \href {https://doi.org/10.18653/v1/P16-2090} {Single-model
  encoder-decoder with explicit morphological representation for reinflection}.
\newblock In \emph{Proceedings of the 54th Annual Meeting of the Association
  for Computational Linguistics (Volume 2: Short Papers)}, pages 555--560,
  Berlin, Germany. Association for Computational Linguistics.

\bibitem[{Kann and Sch{\"u}tze(2017)}]{kann-schutze-2017-unlabeled}
Katharina Kann and Hinrich Sch{\"u}tze. 2017.
\newblock \href {https://doi.org/10.18653/v1/W17-4111} {Unlabeled data for
  morphological generation with character-based sequence-to-sequence models}.
\newblock In \emph{Proceedings of the First Workshop on Subword and Character
  Level Models in {NLP}}, pages 76--81, Copenhagen, Denmark. Association for
  Computational Linguistics.

\bibitem[{Kann and Sch{\"u}tze(2018)}]{kann-schutze-2018-neural}
Katharina Kann and Hinrich Sch{\"u}tze. 2018.
\newblock \href {https://doi.org/10.18653/v1/D18-1363} {Neural transductive
  learning and beyond: Morphological generation in the minimal-resource
  setting}.
\newblock In \emph{Proceedings of the 2018 Conference on Empirical Methods in
  Natural Language Processing}, pages 3254--3264, Brussels, Belgium.
  Association for Computational Linguistics.

\bibitem[{Kementchedjhieva et~al.(2018)Kementchedjhieva, Bjerva, and
  Augenstein}]{kementchedjhieva-etal-2018-copenhagen}
Yova Kementchedjhieva, Johannes Bjerva, and Isabelle Augenstein. 2018.
\newblock \href {https://doi.org/10.18653/v1/K18-3011} {Copenhagen at
  {C}o{NLL}{--}{SIGMORPHON} 2018: Multilingual inflection in context with
  explicit morphosyntactic decoding}.
\newblock In \emph{Proceedings of the {C}o{NLL}{--}{SIGMORPHON} 2018 Shared
  Task: Universal Morphological Reinflection}, pages 93--98, Brussels.
  Association for Computational Linguistics.

\bibitem[{Khalifa et~al.(2018)Khalifa, Habash, Eryani, Obeid, Abdulrahim, and
  Al~Kaabi}]{khalifa2018morphologically}
Salam Khalifa, Nizar Habash, Fadhl Eryani, Ossama Obeid, Dana Abdulrahim, and
  Meera Al~Kaabi. 2018.
\newblock \href {https://aclanthology.org/L18-1607} {A morphologically
  annotated corpus of {E}mirati {A}rabic}.
\newblock In \emph{Proceedings of the Eleventh International Conference on
  Language Resources and Evaluation ({LREC} 2018)}, Miyazaki, Japan. European
  Language Resources Association (ELRA).

\bibitem[{Kingma and Ba(2014)}]{kingma2014adam}
Diederik~P Kingma and Jimmy Ba. 2014.
\newblock Adam: A method for stochastic optimization.
\newblock \emph{arXiv preprint arXiv:1412.6980}.

\bibitem[{Kirov et~al.(2018)Kirov, Cotterell, Sylak-Glassman, Walther,
  Vylomova, Xia, Faruqui, Mielke, McCarthy, K{\"u}bler, Yarowsky, Eisner, and
  Hulden}]{kirov-etal-2018-unimorph}
Christo Kirov, Ryan Cotterell, John Sylak-Glassman, G{\'e}raldine Walther,
  Ekaterina Vylomova, Patrick Xia, Manaal Faruqui, Sabrina~J. Mielke, Arya
  McCarthy, Sandra K{\"u}bler, et~al. 2018.
\newblock \href {https://www.aclweb.org/anthology/L18-1293} {{U}ni{M}orph 2.0:
  {U}niversal {M}orphology}.
\newblock In \emph{Proceedings of the Eleventh International Conference on
  Language Resources and Evaluation ({LREC} 2018)}, Miyazaki, Japan. European
  Language Resources Association (ELRA).

\bibitem[{Kirov et~al.(2016)Kirov, Sylak-Glassman, Que, and
  Yarowsky}]{kirov-etal-2016-large}
Christo Kirov, John Sylak-Glassman, Roger Que, and David Yarowsky. 2016.
\newblock \href {https://www.aclweb.org/anthology/L16-1498} {Very-large scale
  parsing and normalization of {W}iktionary morphological paradigms}.
\newblock In \emph{Proceedings of the Tenth International Conference on
  Language Resources and Evaluation ({LREC}'16)}, pages 3121--3126,
  Portoro{\v{z}}, Slovenia. European Language Resources Association (ELRA).

\bibitem[{Klavans(2018)}]{ws-2018-modeling-polysynthetic}
Judith~L. Klavans, editor. 2018.
\newblock \href {https://www.aclweb.org/anthology/W18-4800} {\emph{Proceedings
  of the Workshop on Computational Modeling of Polysynthetic Languages}}.
  Association for Computational Linguistics, Santa Fe, New Mexico, USA.

\bibitem[{Klyachko et~al.(2020)Klyachko, Sorokin, Krizhanovskaya, Krizhanovsky,
  and Ryazanskaya}]{klyachko2020lowresourceeval}
Elena Klyachko, Alexey Sorokin, Natalia Krizhanovskaya, Andrew Krizhanovsky,
  and Galina Ryazanskaya. 2020.
\newblock {L}ow{R}esource{E}val-2019: a shared task on morphological analysis
  for low-resource languages.
\newblock \emph{arXiv preprint arXiv:2001.11285}.

\bibitem[{Kondratyuk(2019)}]{kondratyuk-2019-cross}
Dan Kondratyuk. 2019.
\newblock \href {https://doi.org/10.18653/v1/W19-4203} {Cross-lingual
  lemmatization and morphology tagging with two-stage multilingual {BERT}
  fine-tuning}.
\newblock In \emph{Proceedings of the 16th Workshop on Computational Research
  in Phonetics, Phonology, and Morphology}, pages 12--18, Florence, Italy.
  Association for Computational Linguistics.

\bibitem[{Kong et~al.(2015)Kong, Dyer, and Smith}]{kong2015segmental}
Lingpeng Kong, Chris Dyer, and Noah~A Smith. 2015.
\newblock Segmental recurrent neural networks.
\newblock \emph{arXiv preprint arXiv:1511.06018}.

\bibitem[{Kurimo et~al.(2010)Kurimo, Virpioja, Turunen, and
  Lagus}]{kurimo-etal-2010-morpho}
Mikko Kurimo, Sami Virpioja, Ville Turunen, and Krista Lagus. 2010.
\newblock \href {https://www.aclweb.org/anthology/W10-2211} {Morpho challenge
  2005-2010: Evaluations and results}.
\newblock In \emph{Proceedings of the 11th Meeting of the {ACL} Special
  Interest Group on Computational Morphology and Phonology}, pages 87--95,
  Uppsala, Sweden. Association for Computational Linguistics.

\bibitem[{Kylli{\"a}inen and
  Silfverberg(2019)}]{kylliainen-silfverberg-2019-ensembles}
Ilmari Kylli{\"a}inen and Miikka Silfverberg. 2019.
\newblock \href {https://www.aclweb.org/anthology/W19-6132} {Ensembles of
  neural morphological inflection models}.
\newblock In \emph{Proceedings of the 22nd Nordic Conference on Computational
  Linguistics}, pages 304--309, Turku, Finland. Link{\"o}ping University
  Electronic Press.

\bibitem[{Lafferty et~al.(2001)Lafferty, McCallum, and
  Pereira}]{lafferty2001conditional}
John Lafferty, Andrew McCallum, and Fernando~CN Pereira. 2001.
\newblock Conditional random fields: {P}robabilistic models for segmenting and
  labeling sequence data.
\newblock In \emph{ICML}.

\bibitem[{Lignos et~al.(2009)Lignos, Chan, Marcus, and Yang}]{lignos2009rule}
Constantine Lignos, Erwin Chan, Mitchell~P Marcus, and Charles Yang. 2009.
\newblock A rule-based unsupervised morphology learning framework.
\newblock In \emph{CLEF (Working Notes)}.

\bibitem[{Liu et~al.(2018)Liu, Subbiah, Wiemerslage, Lilley, and
  Moeller}]{liu-etal-2018-morphological}
Ling Liu, Ilamvazhuthy Subbiah, Adam Wiemerslage, Jonathan Lilley, and Sarah
  Moeller. 2018.
\newblock \href {https://doi.org/10.18653/v1/K18-3010} {Morphological
  reinflection in context: {CU} boulder{'}s submission to
  {C}o{NLL}{--}{SIGMORPHON} 2018 shared task}.
\newblock In \emph{Proceedings of the {C}o{NLL}{--}{SIGMORPHON} 2018 Shared
  Task: Universal Morphological Reinflection}, pages 86--92, Brussels.
  Association for Computational Linguistics.

\bibitem[{MacWhinney(2014)}]{macwhinney2014childes}
Brian MacWhinney. 2014.
\newblock \emph{The {CHILDES} project: Tools for analyzing talk}.
\newblock Psychology Press.

\bibitem[{Mager et~al.(2020)Mager, {\c{C}}etino{\u{g}}lu, and
  Kann}]{mager-etal-2020-tackling}
Manuel Mager, {\"O}zlem {\c{C}}etino{\u{g}}lu, and Katharina Kann. 2020.
\newblock \href {https://doi.org/10.18653/v1/2020.emnlp-main.423} {Tackling the
  low-resource challenge for canonical segmentation}.
\newblock In \emph{Proceedings of the 2020 Conference on Empirical Methods in
  Natural Language Processing (EMNLP)}, pages 5237--5250, Online. Association
  for Computational Linguistics.

\bibitem[{Mager and Kann(2020)}]{mager-kann-2020-ims}
Manuel Mager and Katharina Kann. 2020.
\newblock \href {https://doi.org/10.18653/v1/2020.sigmorphon-1.9} {The
  {IMS}{--}{CUB}oulder system for the {SIGMORPHON} 2020 shared task on
  unsupervised morphological paradigm completion}.
\newblock In \emph{Proceedings of the 17th SIGMORPHON Workshop on Computational
  Research in Phonetics, Phonology, and Morphology}, pages 99--105, Online.
  Association for Computational Linguistics.

\bibitem[{Mager et~al.(2021)Mager, Oncevay, Ebrahimi, Ortega, Rios, Fan,
  Gutierrez-Vasques, Chiruzzo, Gim{\'e}nez-Lugo, Ramos, Meza~Ruiz, Coto-Solano,
  Palmer, Mager-Hois, Chaudhary, Neubig, Vu, and Kann}]{mager2021findings}
Manuel Mager, Arturo Oncevay, Abteen Ebrahimi, John Ortega, Annette Rios,
  Angela Fan, Ximena Gutierrez-Vasques, Luis Chiruzzo, Gustavo
  Gim{\'e}nez-Lugo, Ricardo Ramos, et~al. 2021.
\newblock \href {https://doi.org/10.18653/v1/2021.americasnlp-1.23} {Findings
  of the {A}mericas{NLP} 2021 shared task on open machine translation for
  indigenous languages of the {A}mericas}.
\newblock In \emph{Proceedings of the First Workshop on Natural Language
  Processing for Indigenous Languages of the Americas}, pages 202--217, Online.
  Association for Computational Linguistics.

\bibitem[{Makarov and
  Clematide(2018{\natexlab{a}})}]{makarov-clematide-2018-imitation}
Peter Makarov and Simon Clematide. 2018{\natexlab{a}}.
\newblock \href {https://doi.org/10.18653/v1/D18-1314} {Imitation learning for
  neural morphological string transduction}.
\newblock In \emph{Proceedings of the 2018 Conference on Empirical Methods in
  Natural Language Processing}, pages 2877--2882, Brussels, Belgium.
  Association for Computational Linguistics.

\bibitem[{Makarov and
  Clematide(2018{\natexlab{b}})}]{makarov-clematide-2018-neural}
Peter Makarov and Simon Clematide. 2018{\natexlab{b}}.
\newblock \href {https://www.aclweb.org/anthology/C18-1008} {Neural
  transition-based string transduction for limited-resource setting in
  morphology}.
\newblock In \emph{Proceedings of the 27th International Conference on
  Computational Linguistics}, pages 83--93, Santa Fe, New Mexico, USA.
  Association for Computational Linguistics.

\bibitem[{Makarov and
  Clematide(2018{\natexlab{c}})}]{makarov-clematide-2018-uzh}
Peter Makarov and Simon Clematide. 2018{\natexlab{c}}.
\newblock \href {https://doi.org/10.18653/v1/K18-3008} {{UZH} at
  {C}o{NLL}{--}{SIGMORPHON} 2018 shared task on universal morphological
  reinflection}.
\newblock In \emph{Proceedings of the {C}o{NLL}{--}{SIGMORPHON} 2018 Shared
  Task: Universal Morphological Reinflection}, pages 69--75, Brussels.
  Association for Computational Linguistics.

\bibitem[{Makarov et~al.(2017)Makarov, Ruzsics, and
  Clematide}]{makarov-etal-2017-align}
Peter Makarov, Tatiana Ruzsics, and Simon Clematide. 2017.
\newblock \href {https://doi.org/10.18653/v1/K17-2004} {Align and copy: {UZH}
  at {SIGMORPHON} 2017 shared task for morphological reinflection}.
\newblock In \emph{Proceedings of the {C}o{NLL} {SIGMORPHON} 2017 Shared Task:
  Universal Morphological Reinflection}, pages 49--57, Vancouver. Association
  for Computational Linguistics.

\bibitem[{Malaviya et~al.(2018)Malaviya, Gormley, and
  Neubig}]{malaviya-etal-2018-neural}
Chaitanya Malaviya, Matthew~R. Gormley, and Graham Neubig. 2018.
\newblock \href {https://doi.org/10.18653/v1/P18-1247} {Neural factor graph
  models for cross-lingual morphological tagging}.
\newblock In \emph{Proceedings of the 56th Annual Meeting of the Association
  for Computational Linguistics (Volume 1: Long Papers)}, pages 2653--2663,
  Melbourne, Australia. Association for Computational Linguistics.

\bibitem[{Malaviya et~al.(2019)Malaviya, Wu, and
  Cotterell}]{malaviya-etal-2019-simple}
Chaitanya Malaviya, Shijie Wu, and Ryan Cotterell. 2019.
\newblock \href {https://doi.org/10.18653/v1/N19-1155} {A simple joint model
  for improved contextual neural lemmatization}.
\newblock In \emph{Proceedings of the 2019 Conference of the North {A}merican
  Chapter of the Association for Computational Linguistics: Human Language
  Technologies, Volume 1 (Long and Short Papers)}, pages 1517--1528,
  Minneapolis, Minnesota. Association for Computational Linguistics.

\bibitem[{Maletti and Constant(2011)}]{maletti2011proceedings}
Andreas Maletti and Matthieu Constant. 2011.
\newblock Proceedings of the 9th international workshop on finite state methods
  and natural language processing.
\newblock In \emph{Proceedings of the 9th International Workshop on Finite
  State Methods and Natural Language Processing}.

\bibitem[{Malouf et~al.(2020)Malouf, Ackerman, and
  Semenuks}]{malouf-etal-2020-lexical}
Robert Malouf, Farrell Ackerman, and Arturs Semenuks. 2020.
\newblock \href {https://www.aclweb.org/anthology/2020.scil-1.52} {Lexical
  databases for computational analyses: A linguistic perspective}.
\newblock In \emph{Proceedings of the Society for Computation in Linguistics
  2020}, pages 446--456, New York, New York. Association for Computational
  Linguistics.

\bibitem[{McCarthy et~al.(2020{\natexlab{a}})McCarthy, Kirov, Grella, Nidhi,
  Xia, Gorman, Vylomova, Mielke, Nicolai, Silfverberg, Arkhangelskiy,
  Krizhanovsky, Krizhanovsky, Klyachko, Sorokin, Mansfield, Ern{\v{s}}treits,
  Pinter, Jacobs, Cotterell, Hulden, and
  Yarowsky}]{mccarthy-etal-2020-unimorph}
Arya~D. McCarthy, Christo Kirov, Matteo Grella, Amrit Nidhi, Patrick Xia, Kyle
  Gorman, Ekaterina Vylomova, Sabrina~J. Mielke, Garrett Nicolai, Miikka
  Silfverberg, et~al. 2020{\natexlab{a}}.
\newblock \href {https://www.aclweb.org/anthology/2020.lrec-1.483}
  {{U}ni{M}orph 3.0: {U}niversal {M}orphology}.
\newblock In \emph{Proceedings of the 12th Language Resources and Evaluation
  Conference}, pages 3922--3931, Marseille, France. European Language Resources
  Association.

\bibitem[{McCarthy et~al.(2018)McCarthy, Silfverberg, Cotterell, Hulden, and
  Yarowsky}]{mccarthy-etal-2018-marrying}
Arya~D. McCarthy, Miikka Silfverberg, Ryan Cotterell, Mans Hulden, and David
  Yarowsky. 2018.
\newblock \href {https://doi.org/10.18653/v1/W18-6011} {Marrying {U}niversal
  {D}ependencies and {U}niversal {M}orphology}.
\newblock In \emph{Proceedings of the Second Workshop on Universal Dependencies
  ({UDW} 2018)}, pages 91--101, Brussels, Belgium. Association for
  Computational Linguistics.

\bibitem[{McCarthy et~al.(2019)McCarthy, Vylomova, Wu, Malaviya, Wolf-Sonkin,
  Nicolai, Kirov, Silfverberg, Mielke, Heinz, Cotterell, and
  Hulden}]{mccarthy-etal-2019-sigmorphon}
Arya~D. McCarthy, Ekaterina Vylomova, Shijie Wu, Chaitanya Malaviya, Lawrence
  Wolf-Sonkin, Garrett Nicolai, Christo Kirov, Miikka Silfverberg, Sabrina~J.
  Mielke, Jeffrey Heinz, et~al. 2019.
\newblock \href {https://doi.org/10.18653/v1/W19-4226} {The {SIGMORPHON} 2019
  shared task: Morphological analysis in context and cross-lingual transfer for
  inflection}.
\newblock In \emph{Proceedings of the 16th Workshop on Computational Research
  in Phonetics, Phonology, and Morphology}, pages 229--244, Florence, Italy.
  Association for Computational Linguistics.

\bibitem[{McCarthy et~al.(2020{\natexlab{b}})McCarthy, Wicks, Lewis, Mueller,
  Wu, Adams, Nicolai, Post, and Yarowsky}]{mccarthy-etal-2020-johns}
Arya~D. McCarthy, Rachel Wicks, Dylan Lewis, Aaron Mueller, Winston Wu, Oliver
  Adams, Garrett Nicolai, Matt Post, and David Yarowsky. 2020{\natexlab{b}}.
\newblock \href {https://www.aclweb.org/anthology/2020.lrec-1.352} {The {J}ohns
  {H}opkins {U}niversity {B}ible corpus: 1600+ tongues for typological
  exploration}.
\newblock In \emph{Proceedings of the 12th Language Resources and Evaluation
  Conference}, pages 2884--2892, Marseille, France. European Language Resources
  Association.

\bibitem[{McCurdy et~al.(2021)McCurdy, Goldwater, and
  Lopez}]{mccurdy-etal-2021-adaptor}
Kate McCurdy, Sharon Goldwater, and Adam Lopez. 2021.
\newblock \href {https://doi.org/10.18653/v1/2021.sigmorphon-1.9} {{A}daptor
  {G}rammars for unsupervised paradigm clustering}.
\newblock In \emph{Proceedings of the 18th SIGMORPHON Workshop on Computational
  Research in Phonetics, Phonology, and Morphology}, pages 82--89, Online.
  Association for Computational Linguistics.

\bibitem[{Metheniti and Neumann(2020)}]{metheniti-neumann-2020-wikinflection}
Eleni Metheniti and Guenter Neumann. 2020.
\newblock \href {https://www.aclweb.org/anthology/2020.lrec-1.481}
  {Wikinflection corpus: A (better) multilingual, morpheme-annotated
  inflectional corpus}.
\newblock In \emph{Proceedings of the 12th Language Resources and Evaluation
  Conference}, pages 3905--3912, Marseille, France. European Language Resources
  Association.

\bibitem[{Michel and Neubig(2018)}]{michel-neubig-2018-mtnt}
Paul Michel and Graham Neubig. 2018.
\newblock \href {https://doi.org/10.18653/v1/D18-1050} {{MTNT}: A testbed for
  machine translation of noisy text}.
\newblock In \emph{Proceedings of the 2018 Conference on Empirical Methods in
  Natural Language Processing}, pages 543--553, Brussels, Belgium. Association
  for Computational Linguistics.

\bibitem[{Micher(2017)}]{micher-2017-improving}
Jeffrey Micher. 2017.
\newblock \href {https://doi.org/10.18653/v1/W17-0114} {Improving coverage of
  an {I}nuktitut morphological analyzer using a segmental recurrent neural
  network}.
\newblock In \emph{Proceedings of the 2nd Workshop on the Use of Computational
  Methods in the Study of Endangered Languages}, pages 101--106, Honolulu.
  Association for Computational Linguistics.

\bibitem[{Moeller et~al.(2020)Moeller, Liu, Yang, Kann, and
  Hulden}]{moeller-etal-2020-igt2p}
Sarah Moeller, Ling Liu, Changbing Yang, Katharina Kann, and Mans Hulden. 2020.
\newblock \href {https://doi.org/10.18653/v1/2020.emnlp-main.424} {{IGT}2{P}:
  From interlinear glossed texts to paradigms}.
\newblock In \emph{Proceedings of the 2020 Conference on Empirical Methods in
  Natural Language Processing (EMNLP)}, pages 5251--5262, Online. Association
  for Computational Linguistics.

\bibitem[{Moeng et~al.(2021)Moeng, Reay, Daniels, and
  Buys}]{moeng2021canonical}
Tumi Moeng, Sheldon Reay, Aaron Daniels, and Jan Buys. 2021.
\newblock \href {http://arxiv.org/abs/2104.00767} {Canonical and surface
  morphological segmentation for {N}guni languages}.
\newblock \emph{CoRR}, abs/2104.00767.

\bibitem[{Monson et~al.(2007)Monson, Carbonell, Lavie, and
  Levin}]{monson2007paramor}
Christian Monson, Jaime Carbonell, Alon Lavie, and Lori Levin. 2007.
\newblock {P}ara{M}or: Finding paradigms across morphology.
\newblock In \emph{Workshop of the Cross-Language Evaluation Forum for European
  Languages}, pages 900--907. Springer.

\bibitem[{Montag et~al.(2015)Montag, Jones, and
  Smith}]{doi:10.1177/0956797615594361}
Jessica~L. Montag, Michael~N. Jones, and Linda~B. Smith. 2015.
\newblock \href {https://doi.org/10.1177/0956797615594361} {The words children
  hear: Picture books and the statistics for language learning}.
\newblock \emph{Psychological Science}, 26(9):1489--1496.
\newblock PMID: 26243292.

\bibitem[{Mueller et~al.(2013)Mueller, Schmid, and
  Sch{\"u}tze}]{mueller-etal-2013-efficient}
Thomas Mueller, Helmut Schmid, and Hinrich Sch{\"u}tze. 2013.
\newblock \href {https://www.aclweb.org/anthology/D13-1032} {Efficient
  higher-order {CRF}s for morphological tagging}.
\newblock In \emph{Proceedings of the 2013 Conference on Empirical Methods in
  Natural Language Processing}, pages 322--332, Seattle, Washington, USA.
  Association for Computational Linguistics.

\bibitem[{Narasimhan et~al.(2015)Narasimhan, Barzilay, and
  Jaakkola}]{narasimhan2015unsupervised}
Karthik Narasimhan, Regina Barzilay, and Tommi Jaakkola. 2015.
\newblock \href {https://doi.org/10.1162/tacl_a_00130} {An unsupervised method
  for uncovering morphological chains}.
\newblock \emph{Transactions of the Association for Computational Linguistics},
  3:157--167.

\bibitem[{Nguyen et~al.(2021)Nguyen, Lai, Pouran Ben~Veyseh, and
  Nguyen}]{nguyen-etal-2021-trankit}
Minh~Van Nguyen, Viet~Dac Lai, Amir Pouran Ben~Veyseh, and Thien~Huu Nguyen.
  2021.
\newblock \href {https://doi.org/10.18653/v1/2021.eacl-demos.10} {Trankit: A
  light-weight transformer-based toolkit for multilingual natural language
  processing}.
\newblock In \emph{Proceedings of the 16th Conference of the European Chapter
  of the Association for Computational Linguistics: System Demonstrations},
  pages 80--90, Online. Association for Computational Linguistics.

\bibitem[{Nicolai et~al.(2015)Nicolai, Cherry, and
  Kondrak}]{nicolai-etal-2015-inflection}
Garrett Nicolai, Colin Cherry, and Grzegorz Kondrak. 2015.
\newblock \href {https://doi.org/10.3115/v1/N15-1093} {Inflection generation as
  discriminative string transduction}.
\newblock In \emph{Proceedings of the 2015 Conference of the North {A}merican
  Chapter of the Association for Computational Linguistics: Human Language
  Technologies}, pages 922--931, Denver, Colorado. Association for
  Computational Linguistics.

\bibitem[{Nicolai et~al.(2021)Nicolai, Gorman, and
  Cotterell}]{sigmorphon-2021-sigmorphon}
Garrett Nicolai, Kyle Gorman, and Ryan Cotterell, editors. 2021.
\newblock \href {https://aclanthology.org/2021.sigmorphon-1.0}
  {\emph{Proceedings of the 18th {SIGMORPHON} Workshop on Computational
  Research in Phonetics, Phonology, and Morphology}}. Association for
  Computational Linguistics, Online.

\bibitem[{Nicolai et~al.(2020)Nicolai, Lewis, McCarthy, Mueller, Wu, and
  Yarowsky}]{nicolai-etal-2020-fine}
Garrett Nicolai, Dylan Lewis, Arya~D. McCarthy, Aaron Mueller, Winston Wu, and
  David Yarowsky. 2020.
\newblock \href {https://www.aclweb.org/anthology/2020.lrec-1.488}
  {Fine-grained morphosyntactic analysis and generation tools for more than one
  thousand languages}.
\newblock In \emph{Proceedings of the 12th Language Resources and Evaluation
  Conference}, pages 3963--3972, Marseille, France. European Language Resources
  Association.

\bibitem[{Ott et~al.(2019)Ott, Edunov, Baevski, Fan, Gross, Ng, Grangier, and
  Auli}]{ott-etal-2019-fairseq}
Myle Ott, Sergey Edunov, Alexei Baevski, Angela Fan, Sam Gross, Nathan Ng,
  David Grangier, and Michael Auli. 2019.
\newblock \href {https://doi.org/10.18653/v1/N19-4009} {fairseq: A fast,
  extensible toolkit for sequence modeling}.
\newblock In \emph{Proceedings of the 2019 Conference of the North {A}merican
  Chapter of the Association for Computational Linguistics (Demonstrations)},
  pages 48--53, Minneapolis, Minnesota. Association for Computational
  Linguistics.

\bibitem[{Park et~al.(2021)Park, Zhang, Haley, Steimel, Liu, and
  Schwartz}]{10.1162/tacl_a_00365}
Hyunji~Hayley Park, Katherine~J. Zhang, Coleman Haley, Kenneth Steimel, Han
  Liu, and Lane Schwartz. 2021.
\newblock \href {https://doi.org/10.1162/tacl_a_00365} {{Morphology Matters: A
  Multilingual Language Modeling Analysis}}.
\newblock \emph{Transactions of the Association for Computational Linguistics},
  9:261--276.

\bibitem[{Pimentel et~al.(2019)Pimentel, McCarthy, Blasi, Roark, and
  Cotterell}]{pimentel-etal-2019-meaning}
Tiago Pimentel, Arya~D. McCarthy, Damian Blasi, Brian Roark, and Ryan
  Cotterell. 2019.
\newblock \href {https://doi.org/10.18653/v1/P19-1171} {Meaning to form:
  Measuring systematicity as information}.
\newblock In \emph{Proceedings of the 57th Annual Meeting of the Association
  for Computational Linguistics}, pages 1751--1764, Florence, Italy.
  Association for Computational Linguistics.

\bibitem[{Pimentel et~al.(2021)Pimentel, Ryskina, Mielke, Wu, Chodroff,
  Leonard, Nicolai, Ghanggo~Ate, Khalifa, Habash, El-Khaissi, Goldman, Gasser,
  Lane, Coler, Oncevay, Montoya~Samame, Silva~Villegas, Ek, Bernardy,
  Shcherbakov, Bayyr-ool, Sheifer, Ganieva, Plugaryov, Klyachko, Salehi,
  Krizhanovsky, Krizhanovsky, Vania, Ivanova, Salchak, Straughn, Liu,
  Washington, Ataman, Kiera{\'s}, Woli{\'n}ski, Suhardijanto, Stoehr, Nuriah,
  Ratan, Tyers, Ponti, Aiton, Hatcher, Prud'hommeaux, Kumar, Hulden, Barta,
  Lakatos, Szolnok, {\'A}cs, Raj, Yarowsky, Cotterell, Ambridge, and
  Vylomova}]{pimentel-ryskina-etal-2021-sigmorphon}
Tiago Pimentel, Maria Ryskina, Sabrina~J. Mielke, Shijie Wu, Eleanor Chodroff,
  Brian Leonard, Garrett Nicolai, Yustinus Ghanggo~Ate, Salam Khalifa, Nizar
  Habash, et~al. 2021.
\newblock \href {https://doi.org/10.18653/v1/2021.sigmorphon-1.25} {Sigmorphon
  2021 shared task on morphological reinflection: Generalization across
  languages}.
\newblock In \emph{Proceedings of the 18th SIGMORPHON Workshop on Computational
  Research in Phonetics, Phonology, and Morphology}, pages 229--259, Online.
  Association for Computational Linguistics.

\bibitem[{Poon et~al.(2009)Poon, Cherry, and Toutanova}]{poon2009unsupervised}
Hoifung Poon, Colin Cherry, and Kristina Toutanova. 2009.
\newblock \href {https://aclanthology.org/N09-1024} {Unsupervised morphological
  segmentation with log-linear models}.
\newblock In \emph{Proceedings of Human Language Technologies: The 2009 Annual
  Conference of the North {A}merican Chapter of the Association for
  Computational Linguistics}, pages 209--217, Boulder, Colorado. Association
  for Computational Linguistics.

\bibitem[{Qi et~al.(2020)Qi, Zhang, Zhang, Bolton, and Manning}]{qi2020stanza}
Peng Qi, Yuhao Zhang, Yuhui Zhang, Jason Bolton, and Christopher~D. Manning.
  2020.
\newblock \href {https://doi.org/10.18653/v1/2020.acl-demos.14} {{S}tanza: A
  python natural language processing toolkit for many human languages}.
\newblock In \emph{Proceedings of the 58th Annual Meeting of the Association
  for Computational Linguistics: System Demonstrations}, pages 101--108,
  Online. Association for Computational Linguistics.

\bibitem[{Rasooli and Collins(2017)}]{rasooli-collins-2017-cross}
Mohammad~Sadegh Rasooli and Michael Collins. 2017.
\newblock \href {https://doi.org/10.1162/tacl_a_00061} {Cross-lingual syntactic
  transfer with limited resources}.
\newblock \emph{Transactions of the Association for Computational Linguistics},
  5:279--293.

\bibitem[{Ruokolainen et~al.(2013)Ruokolainen, Kohonen, Virpioja, and
  Kurimo}]{ruokolainen-etal-2013-supervised}
Teemu Ruokolainen, Oskar Kohonen, Sami Virpioja, and Mikko Kurimo. 2013.
\newblock \href {https://www.aclweb.org/anthology/W13-3504} {Supervised
  morphological segmentation in a low-resource learning setting using
  conditional random fields}.
\newblock In \emph{Proceedings of the Seventeenth Conference on Computational
  Natural Language Learning}, pages 29--37, Sofia, Bulgaria. Association for
  Computational Linguistics.

\bibitem[{Ruokolainen et~al.(2014)Ruokolainen, Kohonen, Virpioja, and
  Kurimo}]{ruokolainen-etal-2014-painless}
Teemu Ruokolainen, Oskar Kohonen, Sami Virpioja, and Mikko Kurimo. 2014.
\newblock \href {https://doi.org/10.3115/v1/E14-4017} {Painless semi-supervised
  morphological segmentation using conditional random fields}.
\newblock In \emph{Proceedings of the 14th Conference of the {E}uropean Chapter
  of the Association for Computational Linguistics, volume 2: Short Papers},
  pages 84--89, Gothenburg, Sweden. Association for Computational Linguistics.

\bibitem[{Ruzsics and
  Samard{\v{z}}i{\'c}(2017)}]{ruzsics-samardzic-2017-neural}
Tatyana Ruzsics and Tanja Samard{\v{z}}i{\'c}. 2017.
\newblock \href {https://doi.org/10.18653/v1/K17-1020} {Neural
  sequence-to-sequence learning of internal word structure}.
\newblock In \emph{Proceedings of the 21st Conference on Computational Natural
  Language Learning ({C}o{NLL} 2017)}, pages 184--194, Vancouver, Canada.
  Association for Computational Linguistics.

\bibitem[{Schone and Jurafsky(2001)}]{schone-jurafsky-2001-knowledge}
Patrick Schone and Daniel Jurafsky. 2001.
\newblock \href {https://www.aclweb.org/anthology/N01-1024} {Knowledge-free
  induction of inflectional morphologies}.
\newblock In \emph{Second Meeting of the North {A}merican Chapter of the
  Association for Computational Linguistics}.

\bibitem[{Seeker and {\c{C}}etino{\u{g}}lu(2015)}]{seeker-cetinoglu-2015-graph}
Wolfgang Seeker and {\"O}zlem {\c{C}}etino{\u{g}}lu. 2015.
\newblock \href {https://doi.org/10.1162/tacl_a_00144} {A graph-based lattice
  dependency parser for joint morphological segmentation and syntactic
  analysis}.
\newblock \emph{Transactions of the Association for Computational Linguistics},
  3:359--373.

\bibitem[{Seker and Tsarfaty(2020)}]{seker-tsarfaty-2020-pointer}
Amit Seker and Reut Tsarfaty. 2020.
\newblock \href {https://doi.org/10.18653/v1/2020.findings-emnlp.391} {A
  pointer network architecture for joint morphological segmentation and
  tagging}.
\newblock In \emph{Findings of the Association for Computational Linguistics:
  EMNLP 2020}, pages 4368--4378, Online. Association for Computational
  Linguistics.

\bibitem[{Sharma et~al.(2018)Sharma, Katrapati, and
  Sharma}]{sharma-etal-2018-iit}
Abhishek Sharma, Ganesh Katrapati, and Dipti~Misra Sharma. 2018.
\newblock \href {https://doi.org/10.18653/v1/K18-3013} {{IIT}({BHU}){--}{IIITH}
  at {C}o{NLL}{--}{SIGMORPHON} 2018 shared task on universal morphological
  reinflection}.
\newblock In \emph{Proceedings of the {C}o{NLL}{--}{SIGMORPHON} 2018 Shared
  Task: Universal Morphological Reinflection}, pages 105--111, Brussels.
  Association for Computational Linguistics.

\bibitem[{Silfverberg and Hulden(2018)}]{silfverberg-hulden-2018-encoder}
Miikka Silfverberg and Mans Hulden. 2018.
\newblock \href {https://doi.org/10.18653/v1/D18-1315} {An encoder-decoder
  approach to the paradigm cell filling problem}.
\newblock In \emph{Proceedings of the 2018 Conference on Empirical Methods in
  Natural Language Processing}, pages 2883--2889, Brussels, Belgium.
  Association for Computational Linguistics.

\bibitem[{Silfverberg et~al.(2017)Silfverberg, Wiemerslage, Liu, and
  Mao}]{silfverberg-etal-2017-data}
Miikka Silfverberg, Adam Wiemerslage, Ling Liu, and Lingshuang~Jack Mao. 2017.
\newblock \href {https://doi.org/10.18653/v1/K17-2010} {Data augmentation for
  morphological reinflection}.
\newblock In \emph{Proceedings of the {C}o{NLL} {SIGMORPHON} 2017 Shared Task:
  Universal Morphological Reinflection}, pages 90--99, Vancouver. Association
  for Computational Linguistics.

\bibitem[{Singer and Kann(2020)}]{singer-kann-2020-nyu}
Assaf Singer and Katharina Kann. 2020.
\newblock \href {https://doi.org/10.18653/v1/2020.sigmorphon-1.8} {The
  {NYU}-{CUB}oulder systems for {SIGMORPHON} 2020 task 0 and task 2}.
\newblock In \emph{Proceedings of the 17th SIGMORPHON Workshop on Computational
  Research in Phonetics, Phonology, and Morphology}, pages 90--98, Online.
  Association for Computational Linguistics.

\bibitem[{Smit et~al.(2014)Smit, Virpioja, Gr{\"o}nroos, Kurimo
  et~al.}]{smit2014morfessor}
Peter Smit, Sami Virpioja, Stig-Arne Gr{\"o}nroos, Mikko Kurimo, et~al. 2014.
\newblock Morfessor 2.0: Toolkit for statistical morphological segmentation.
\newblock In \emph{The 14th Conference of the European Chapter of the
  Association for Computational Linguistics (EACL), Gothenburg, Sweden, April
  26-30, 2014}. Aalto University.

\bibitem[{Smith and Eisner(2005)}]{smith2005contrastive}
Noah~A. Smith and Jason Eisner. 2005.
\newblock \href {https://doi.org/10.3115/1219840.1219884} {Contrastive
  estimation: Training log-linear models on unlabeled data}.
\newblock In \emph{Proceedings of the 43rd Annual Meeting of the Association
  for Computational Linguistics ({ACL}{'}05)}, pages 354--362, Ann Arbor,
  Michigan. Association for Computational Linguistics.

\bibitem[{Sorokin(2019)}]{sorokin-2019-convolutional}
Alexey Sorokin. 2019.
\newblock \href {https://doi.org/10.18653/v1/W19-4218} {Convolutional neural
  networks for low-resource morpheme segmentation: baseline or
  state-of-the-art?}
\newblock In \emph{Proceedings of the 16th Workshop on Computational Research
  in Phonetics, Phonology, and Morphology}, pages 154--159, Florence, Italy.
  Association for Computational Linguistics.

\bibitem[{Straka et~al.(2019)Straka, Strakov{\'a}, and
  Hajic}]{straka-etal-2019-udpipe}
Milan Straka, Jana Strakov{\'a}, and Jan Hajic. 2019.
\newblock \href {https://doi.org/10.18653/v1/W19-4212} {{UDP}ipe at
  {SIGMORPHON} 2019: Contextualized embeddings, regularization with
  morphological categories, corpora merging}.
\newblock In \emph{Proceedings of the 16th Workshop on Computational Research
  in Phonetics, Phonology, and Morphology}, pages 95--103, Florence, Italy.
  Association for Computational Linguistics.

\bibitem[{Sylak-Glassman et~al.(2015{\natexlab{a}})Sylak-Glassman, Kirov, Post,
  Que, and Yarowsky}]{10.1007/978-3-319-23980-4_5}
John Sylak-Glassman, Christo Kirov, Matt Post, Roger Que, and David Yarowsky.
  2015{\natexlab{a}}.
\newblock A universal feature schema for rich morphological annotation and
  fine-grained cross-lingual part-of-speech tagging.
\newblock In \emph{Systems and Frameworks for Computational Morphology}, pages
  72--93, Cham. Springer International Publishing.

\bibitem[{Sylak-Glassman et~al.(2015{\natexlab{b}})Sylak-Glassman, Kirov,
  Yarowsky, and Que}]{sylak-glassman-etal-2015-language}
John Sylak-Glassman, Christo Kirov, David Yarowsky, and Roger Que.
  2015{\natexlab{b}}.
\newblock \href {https://doi.org/10.3115/v1/P15-2111} {A language-independent
  feature schema for inflectional morphology}.
\newblock In \emph{Proceedings of the 53rd Annual Meeting of the Association
  for Computational Linguistics and the 7th International Joint Conference on
  Natural Language Processing (Volume 2: Short Papers)}, pages 674--680,
  Beijing, China. Association for Computational Linguistics.

\bibitem[{Szolnok et~al.(2021)Szolnok, Barta, Lakatos, and
  {\'A}cs}]{szolnok-barta-lakatos-etal-2021-bme}
G{\'a}bor Szolnok, Botond Barta, Dorina Lakatos, and Judit {\'A}cs. 2021.
\newblock \href {https://doi.org/10.18653/v1/2021.sigmorphon-1.27} {Bme
  submission for sigmorphon 2021 shared task 0. a three step training approach
  with data augmentation for morphological inflection}.
\newblock In \emph{Proceedings of the 18th SIGMORPHON Workshop on Computational
  Research in Phonetics, Phonology, and Morphology}, pages 268--273, Online.
  Association for Computational Linguistics.

\bibitem[{Tamchyna et~al.(2017)Tamchyna, Weller-Di~Marco, and
  Fraser}]{tamchyna-etal-2017-modeling}
Ale{\v{s}} Tamchyna, Marion Weller-Di~Marco, and Alexander Fraser. 2017.
\newblock \href {https://doi.org/10.18653/v1/W17-4704} {Modeling target-side
  inflection in neural machine translation}.
\newblock In \emph{Proceedings of the Second Conference on Machine
  Translation}, pages 32--42, Copenhagen, Denmark. Association for
  Computational Linguistics.

\bibitem[{Vaswani et~al.(2017)Vaswani, Shazeer, Parmar, Uszkoreit, Jones,
  Gomez, Kaiser, and Polosukhin}]{vaswani2017attention}
Ashish Vaswani, Noam Shazeer, Niki Parmar, Jakob Uszkoreit, Llion Jones,
  Aidan~N Gomez, {\L}ukasz Kaiser, and Illia Polosukhin. 2017.
\newblock Attention is all you need.
\newblock \emph{Advances in neural information processing systems}, 30.

\bibitem[{Vinyals et~al.(2015)Vinyals, Fortunato, and
  Jaitly}]{vinyals2015pointer}
Oriol Vinyals, Meire Fortunato, and Navdeep Jaitly. 2015.
\newblock Pointer networks.
\newblock \emph{Advances in neural information processing systems}, 28.

\bibitem[{Virpioja et~al.(2009)Virpioja, Kohonen, and
  Lagus}]{virpioja2009unsupervised}
Sami Virpioja, Oskar Kohonen, and Krista Lagus. 2009.
\newblock Unsupervised morpheme discovery with {A}llomorfessor.
\newblock In \emph{CLEF (Working Notes)}.

\bibitem[{Wang et~al.(2016)Wang, Cao, Xia, and De~Melo}]{wang2016morphological}
Linlin Wang, Zhu Cao, Yu~Xia, and Gerard De~Melo. 2016.
\newblock Morphological segmentation with window {LSTM} neural networks.
\newblock In \emph{Thirtieth AAAI Conference on Artificial Intelligence}.

\bibitem[{Wang et~al.(2019)Wang, Fam, Bao, Lepage, and Gao}]{wang2019neural}
Weihua Wang, Rashel Fam, Feilong Bao, Yves Lepage, and Guanglai Gao. 2019.
\newblock Neural morphological segmentation model for {M}ongolian.
\newblock In \emph{2019 International Joint Conference on Neural Networks
  (IJCNN)}, pages 1--7. IEEE.

\bibitem[{Wiemerslage et~al.(2021)Wiemerslage, McCarthy, Erdmann, Nicolai,
  Agirrezabal, Silfverberg, Hulden, and Kann}]{wiemerslage-etal-2021-findings}
Adam Wiemerslage, Arya~D. McCarthy, Alexander Erdmann, Garrett Nicolai, Manex
  Agirrezabal, Miikka Silfverberg, Mans Hulden, and Katharina Kann. 2021.
\newblock \href {https://doi.org/10.18653/v1/2021.sigmorphon-1.8} {Findings of
  the {SIGMORPHON} 2021 shared task on unsupervised morphological paradigm
  clustering}.
\newblock In \emph{Proceedings of the 18th SIGMORPHON Workshop on Computational
  Research in Phonetics, Phonology, and Morphology}, pages 72--81, Online.
  Association for Computational Linguistics.

\bibitem[{Wu et~al.(2021)Wu, Cotterell, and Hulden}]{wu-etal-2021-applying}
Shijie Wu, Ryan Cotterell, and Mans Hulden. 2021.
\newblock \href {https://doi.org/10.18653/v1/2021.eacl-main.163} {Applying the
  transformer to character-level transduction}.
\newblock In \emph{Proceedings of the 16th Conference of the European Chapter
  of the Association for Computational Linguistics: Main Volume}, pages
  1901--1907, Online. Association for Computational Linguistics.

\bibitem[{Wu et~al.(2019)Wu, Cotterell, and
  O{'}Donnell}]{wu-etal-2019-morphological}
Shijie Wu, Ryan Cotterell, and Timothy O{'}Donnell. 2019.
\newblock \href {https://doi.org/10.18653/v1/P19-1505} {Morphological
  irregularity correlates with frequency}.
\newblock In \emph{Proceedings of the 57th Annual Meeting of the Association
  for Computational Linguistics}, pages 5117--5126, Florence, Italy.
  Association for Computational Linguistics.

\bibitem[{Xu et~al.(2020)Xu, Kodner, Marcus, and Yang}]{xu-etal-2020-modeling}
Hongzhi Xu, Jordan Kodner, Mitchell Marcus, and Charles Yang. 2020.
\newblock \href {https://doi.org/10.18653/v1/2020.acl-main.596} {Modeling
  morphological typology for unsupervised learning of language morphology}.
\newblock In \emph{Proceedings of the 58th Annual Meeting of the Association
  for Computational Linguistics}, pages 6672--6681, Online. Association for
  Computational Linguistics.

\bibitem[{Xu et~al.(2018)Xu, Yang, Otani, and Wu}]{xu-etal-2018-unsupervised}
Ruochen Xu, Yiming Yang, Naoki Otani, and Yuexin Wu. 2018.
\newblock \href {https://doi.org/10.18653/v1/D18-1268} {Unsupervised
  cross-lingual transfer of word embedding spaces}.
\newblock In \emph{Proceedings of the 2018 Conference on Empirical Methods in
  Natural Language Processing}, pages 2465--2474, Brussels, Belgium.
  Association for Computational Linguistics.

\bibitem[{Yang et~al.(2021)Yang, Nicolai, and
  Silfverberg}]{yang-etal-2021-unsupervised}
Changbing Yang, Garrett Nicolai, and Miikka Silfverberg. 2021.
\newblock \href {https://doi.org/10.18653/v1/2021.sigmorphon-1.11}
  {Unsupervised paradigm clustering using transformation rules}.
\newblock In \emph{Proceedings of the 18th SIGMORPHON Workshop on Computational
  Research in Phonetics, Phonology, and Morphology}, pages 98--106, Online.
  Association for Computational Linguistics.

\bibitem[{Yarowsky et~al.(2001)Yarowsky, Ngai, and
  Wicentowski}]{yarowsky-etal-2001-inducing}
David Yarowsky, Grace Ngai, and Richard Wicentowski. 2001.
\newblock \href {https://www.aclweb.org/anthology/H01-1035} {Inducing
  multilingual text analysis tools via robust projection across aligned
  corpora}.
\newblock In \emph{Proceedings of the First International Conference on Human
  Language Technology Research}.

\bibitem[{Yarowsky and Wicentowski(2000)}]{yarowsky-wicentowski-2000-minimally}
David Yarowsky and Richard Wicentowski. 2000.
\newblock \href {https://doi.org/10.3115/1075218.1075245} {Minimally supervised
  morphological analysis by multimodal alignment}.
\newblock In \emph{Proceedings of the 38th Annual Meeting of the Association
  for Computational Linguistics}, pages 207--216, Hong Kong. Association for
  Computational Linguistics.

\bibitem[{Zalmout and Habash(2020)}]{zalmout-habash-2020-utilizing}
Nasser Zalmout and Nizar Habash. 2020.
\newblock \href {https://www.aclweb.org/anthology/2020.coling-main.412}
  {Utilizing subword entities in character-level sequence-to-sequence
  lemmatization models}.
\newblock In \emph{Proceedings of the 28th International Conference on
  Computational Linguistics}, pages 4676--4682, Barcelona, Spain (Online).
  International Committee on Computational Linguistics.

\bibitem[{Zeman et~al.(2021)Zeman, Nivre, Abrams, Ackermann, Aepli, Aghaei,
  Agi{\'c}, Ahmadi, Ahrenberg, Ajede, Aleksandravi{\v c}i{\=u}t{\.e}, Alfina,
  Antonsen, Aplonova, Aquino, Aragon, Aranzabe, Ar{\i}can, Arnard{\'o}ttir,
  Arutie, Arwidarasti, Asahara, Aslan, Ateyah, Atmaca, Attia, Atutxa,
  Augustinus, Badmaeva, Balasubramani, Ballesteros, Banerjee, Bank,
  Barbu~Mititelu, Barkarson, Basmov, Batchelor, Bauer, Bedir, Bengoetxea, Berk,
  Berzak, Bhat, Bhat, Biagetti, Bick, Bielinskien{\.e}, Bjarnad{\'o}ttir,
  Blokland, Bobicev, Boizou, Borges~V{\"o}lker, B{\"o}rstell, Bosco, Bouma,
  Bowman, Boyd, Braggaar, Brokait{\.e}, Burchardt, Candito, Caron, Caron,
  Cassidy, Cavalcanti, Cebiro{\u g}lu~Eryi{\u g}it, Cecchini, Celano, {\v
  C}{\'e}pl{\"o}, Cesur, Cetin, {\c C}etino{\u g}lu, Chalub, Chauhan, Chi,
  Chika, Cho, Choi, Chun, Cignarella, Cinkov{\'a}, Collomb, {\c C}{\"o}ltekin,
  Connor, Courtin, Cristescu, Daniel, Davidson, de~Marneffe, de~Paiva, Derin,
  de~Souza, Diaz~de Ilarraza, Dickerson, Dinakaramani, Di~Nuovo, Dione, Dirix,
  Dobrovoljc, Dozat, Droganova, Dwivedi, Eckhoff, Eiche, Eli, Elkahky, Ephrem,
  Erina, Erjavec, Etienne, Evelyn, Facundes, Farkas, Fernanda,
  Fernandez~Alcalde, Foster, Freitas, Fujita, Gajdo{\v s}ov{\'a}, Galbraith,
  Garcia, G{\"a}rdenfors, Garza, Gerardi, Gerdes, Ginter, Godoy, Goenaga,
  Gojenola, G{\"o}k{\i}rmak, Goldberg, G{\'o}mez~Guinovart,
  Gonz{\'a}lez~Saavedra, Grici{\=u}t{\.e}, Grioni, Grobol, Gr{\=
  u}z{\={\i}}tis, Guillaume, Guillot-Barbance, G{\"u}ng{\"o}r, Habash,
  Hafsteinsson, Haji{\v c}, Haji{\v c}~jr., H{\"a}m{\"a}l{\"a}inen,
  H{\`a}~M{\~y}, Han, Hanifmuti, Hardwick, Harris, Haug, Heinecke, Hellwig,
  Hennig, Hladk{\'a}, Hlav{\'a}{\v c}ov{\'a}, Hociung, Hohle, Huber, Hwang,
  Ikeda, Ingason, Ion, Irimia, Ishola, Ito, Jel{\'{\i}}nek, Jha, Johannsen,
  J{\'o}nsd{\'o}ttir, J{\o}rgensen, Juutinen, K, Ka{\c s}{\i}kara, Kaasen,
  Kabaeva, Kahane, Kanayama, Kanerva, Kara, Katz, Kayadelen, Kenney,
  Kettnerov{\'a}, Kirchner, Klementieva, K{\"o}hn, K{\"o}ksal, Kopacewicz,
  Korkiakangas, Kotsyba, Kovalevskait{\.e}, Krek, Krishnamurthy, Kuyruk{\c c}u,
  Kuzgun, Kwak, Laippala, Lam, Lambertino, Lando, Larasati, Lavrentiev, Lee,
  L{\^e}~H{\`{\^o}}ng, Lenci, Lertpradit, Leung, Levina, Li, Li, Li, Li, Lim,
  Lima~Padovani, Lind{\'e}n, Ljube{\v s}i{\'c}, Loginova, Luthfi, Luukko,
  Lyashevskaya, Lynn, Macketanz, Makazhanov, Mandl, Manning, Manurung, Mar{\c
  s}an, M{\u a}r{\u a}nduc, Mare{\v c}ek, Marheinecke, Mart{\'{\i}}nez~Alonso,
  Martins, Ma{\v s}ek, Matsuda, Matsumoto, Mazzei, {McDonald}, {McGuinness},
  Mendon{\c c}a, Miekka, Mischenkova, Misirpashayeva, Missil{\"a}, Mititelu,
  Mitrofan, Miyao, Mojiri~Foroushani, Moln{\'a}r, Moloodi, Montemagni, More,
  Moreno~Romero, Moretti, Mori, Mori, Morioka, Moro, Mortensen, Moskalevskyi,
  Muischnek, Munro, Murawaki, M{\"u}{\"u}risep, Nainwani, Nakhl{\'e},
  Navarro~Hor{\~n}iacek, Nedoluzhko, Ne{\v s}pore-B{\=e}rzkalne, Nevaci,
  Nguy{\~{\^e}}n~Th{\d i}, Nguy{\~{\^e}}n Th{\d i}~Minh, Nikaido, Nikolaev,
  Nitisaroj, Nourian, Nurmi, Ojala, Ojha, Ol{\'u}{\`o}kun, Omura, Onwuegbuzia,
  Osenova, {\"O}stling, {\O}vrelid, {\"O}zate{\c s}, {\"O}z{\c c}elik,
  {\"O}zg{\"u}r, {\"O}zt{\"u}rk~Ba{\c s}aran, Park, Partanen, Pascual,
  Passarotti, Patejuk, Paulino-Passos, Peljak-{\L}api{\'n}ska, Peng, Perez,
  Perkova, Perrier, Petrov, Petrova, Phelan, Piitulainen, Pirinen, Pitler,
  Plank, Poibeau, Ponomareva, Popel, Pretkalni{\c n}a, Pr{\'e}vost, Prokopidis,
  Przepi{\'o}rkowski, Puolakainen, Pyysalo, Qi, R{\"a}{\"a}bis, Rademaker,
  Rama, Ramasamy, Ramisch, Rashel, Rasooli, Ravishankar, Real, Rebeja, Reddy,
  Rehm, Riabov, Rie{\ss}ler, Rimkut{\.e}, Rinaldi, Rituma, Rocha,
  R{\"o}gnvaldsson, Romanenko, Rosa, Roșca, Rovati, Rudina, Rueter,
  R{\'u}narsson, Sadde, Safari, Sagot, Sahala, Saleh, Salomoni, Samard{\v
  z}i{\'c}, Samson, Sanguinetti, San{\i}yar, S{\"a}rg, Saul{\={\i}}te,
  Sawanakunanon, Saxena, Scannell, Scarlata, Schneider, Schuster, Schwartz,
  Seddah, Seeker, Seraji, Shen, Shimada, Shirasu, Shishkina, Shohibussirri,
  Sichinava, Siewert, Sigurðsson, Silveira, Silveira, Simi, Simionescu,
  Simk{\'o}, {\v S}imkov{\'a}, Simov, Skachedubova, Smith, Soares-Bastos,
  Spadine, Sprugnoli, Steingr{\'{\i}}msson, Stella, Straka, Strickland,
  Strnadov{\'a}, Suhr, Sulestio, Sulubacak, Suzuki, Sz{\'a}nt{\'o}, Taji,
  Takahashi, Tamburini, Tan, Tanaka, Tella, Tellier, Testori, Thomas, Torga,
  Toska, Trosterud, Trukhina, Tsarfaty, T{\"u}rk, Tyers, Uematsu, Untilov,
  Ure{\v s}ov{\'a}, Uria, Uszkoreit, Utka, Vajjala, van~der Goot, Vanhove, van
  Niekerk, van Noord, Varga, Villemonte de~la Clergerie, Vincze, Vlasova,
  Wakasa, Wallenberg, Wallin, Walsh, Wang, Washington, Wendt, Widmer, Williams,
  Wir{\'e}n, Wittern, Woldemariam, Wong, Wr{\'o}blewska, Yako, Yamashita,
  Yamazaki, Yan, Yasuoka, Yavrumyan, Yenice, Y{\i}ld{\i}z, Yu, {\v
  Z}abokrtsk{\'y}, Zahra, Zeldes, Zhu, Zhuravleva, and Ziane}]{11234/1-3683}
Daniel Zeman, Joakim Nivre, Mitchell Abrams, Elia Ackermann, No{\"e}mi Aepli,
  Hamid Aghaei, {\v Z}eljko Agi{\'c}, Amir Ahmadi, Lars Ahrenberg,
  Chika~Kennedy Ajede, et~al. 2021.
\newblock \href {http://hdl.handle.net/11234/1-3683} {Universal dependencies
  2.8}.
\newblock {LINDAT}/{CLARIAH}-{CZ} digital library at the Institute of Formal
  and Applied Linguistics ({{\'U}FAL}), Faculty of Mathematics and Physics,
  Charles University.

\bibitem[{Zueva et~al.(2020)Zueva, Kuznetsova, and
  Tyers}]{zueva-etal-2020-finite}
Anna Zueva, Anastasia Kuznetsova, and Francis Tyers. 2020.
\newblock \href {https://www.aclweb.org/anthology/2020.lrec-1.314} {A
  finite-state morphological analyser for {E}venki}.
\newblock In \emph{Proceedings of the 12th Language Resources and Evaluation
  Conference}, pages 2581--2589, Marseille, France. European Language Resources
  Association.

\end{thebibliography}

\clearpage
\appendix
\counterwithin{figure}{section}  
\counterwithin{table}{section}  

\section{Remaining Results from Main Text}

\begin{figure*}[t!]
\centering
\small
  \includegraphics[width=0.9\linewidth,
  keepaspectratio]{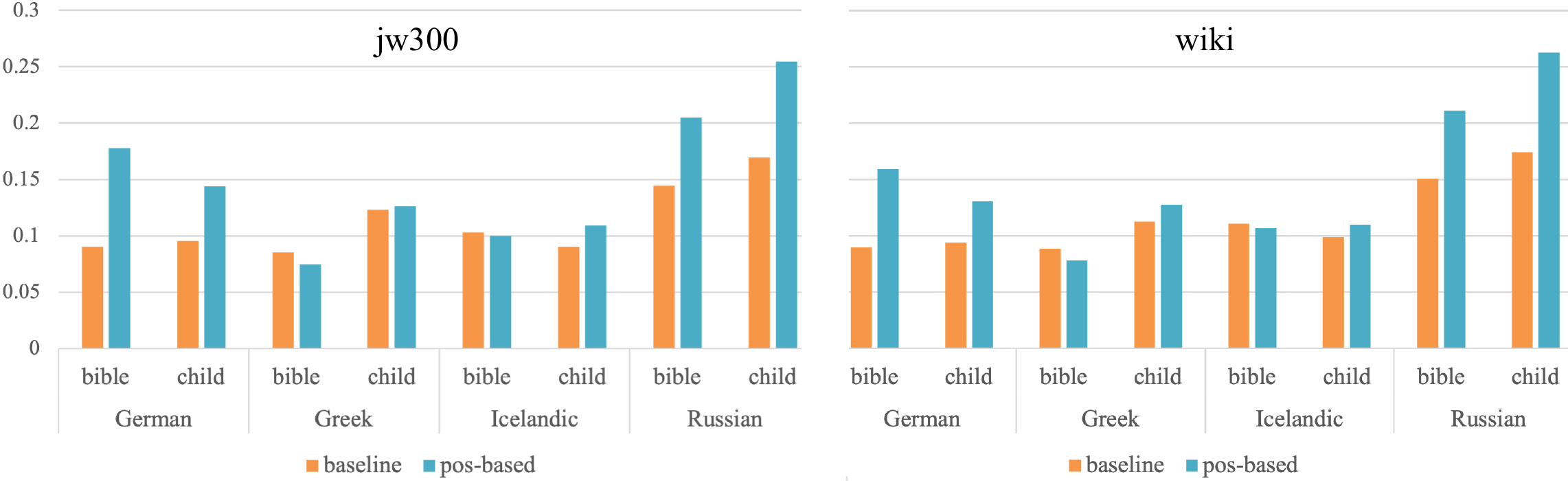}
  \caption{BMAcc for both slot alignment systems on each test corpus, averaged over results for all input clusters. The POS-based system is also averaged over each inflection system.}
  \label{fig:f1_by_slot_sys}
\end{figure*}

The statistics of the data used in our experiments is given in \cref{tab:training_corpora_stats}. Paradigm clustering BMF1 is given in \cref{tab:cluster_results}. Additionally, BMAcc on the two test corpora is given in \cref{fig:f1_by_slot_sys}.

\begin{table*}[t!]
\centering
\small
\setlength{\tabcolsep}{7pt}
\begin{tabular}{l l rrrr}
\toprule
Corpus & Language & Lines & Tokens & Types & \makecell{Type--Token Ratio} \\
\midrule
\multirow[t]{4}{*}{\texttt{Bible}} & German & 31102 & 813317 & 20644 & 0.025 \\
& Greek & 7914 & 194135 & 15541 & 0.080 \\
& Icelandic & 7860 & 185995 & 13050 & 0.070 \\
& Russian & 31102 & 714828 & 43542 & 0.061 \\
\midrule
\multirow[t]{4}{*}{\texttt{Child Directed}} & German & 26592 & 633229 & 31384 & 0.050 \\
& Greek & 8513 & 196344 & 18424 & 0.090 \\
& Icelandic & 8380 & 181687 & 17767 & 0.101 \\
& Russian & 26592 & 586274 & 44823 & 0.077 \\
\bottomrule
\end{tabular}
\caption{\label{tab:training_corpora_stats} Statistics for raw text corpora used for morphology learning}
\end{table*}

\begin{table*}[t]
\centering
\small
\setlength{\tabcolsep}{7pt}
\begin{tabular}{l rrrrr rrrrr}
\toprule
& \multicolumn{5}{c}{Bible} & \multicolumn{5}{c}{Child-Directed} \\
\cmidrule(lr){2-6} \cmidrule{7-11}
System & \lang{Deu} & \lang{Ell} & \lang{Isl} & \lang{Rus} & Average & \lang{Deu} & \lang{Ell} & \lang{Isl} & \lang{Rus} & Average \\
\midrule
\texttt{McC} & \textbf{79.19} & \textbf{81.91} & \textbf{81.66} & \textbf{82.01} & \textbf{81.19} & \textbf{87.72} & \textbf{73.68} & \textbf{84.65} & \textbf{86.28}  & \textbf{83.08} \\
\texttt{Xu} & 63.90 & 65.14 & 67.81 & 52.80 & 63.91 &  70.02 & 46.14 & 55.22 & 63.48 & 58.72 \\
\texttt{SIG} & 46.04 & 57.22 & 47.24 & 45.10 & 48.90 & 45.69 & 47.04 & 43.08 & 47.80 & 45.90 \\
\bottomrule
\end{tabular}
\caption{\label{tab:cluster_results} Paradigm clustering BMF1 scores for a sample of clusters attested in UniMorph.}
\end{table*}

\section{Hyperparameters} \label{app:hyperparams}

\subsection{Morphological Inflection}
\paragraph{Training}
We train all inflection models on the (word, source slot, target slot) triples produced by the slot alignment. Each inflection system considers the word as an input form, and the slots as the tags. We take the hyperparameters from \cite{makarov-clematide-2018-imitation}, and \cite{wu-etal-2021-applying} exactly for each language. For the LSTM, we train a single layer bidirectional encoder with embedding size 100, and LSTM hidden size of 100. The decoder is also a single layer LSTM with hidden size 100. We employ a soft-attention mechanism \cite{bahdanau-et-al}, and optimize with Adam \cite{kingma2014adam} with a learning rate of 0.001, and a gradient clip of 1.0. We train for up to 30 epochs, and a batch size of 16. We employ a soft attention mechanism \cite{bahdanau-et-al}.

\subsection{Slot Prediction}
The slot prediction model is a character-level Transformer encoder-decoder, where both the encoder and decoder have 3 layers and 4 attention heads. We optimize with Adam with a learning rate of 0.0001, and a clip norm of 0.2 for up to 5 epochs.

\section{Additional Details Regarding our Datasets}

\subsection{Statistics of Our Raw-text Corpora}
We give dataset statistics in \autoref{tab:training_corpora_stats}, including type--token ratios.
Bible sizes vary depending on whether or not the Old Testament is included. In the case of smaller Bibles, we down-sample the child-directed corpus to have a roughly equal number of tokens.

\subsection{Test Set Creation} \label{app:test-set-creation}
 We use lemmas and POS tag annotations to match words from the test corpora with UniMorph entries. We sample sentences from the annotated Wikipedia corpora \cite{11234/1-1989} from the ConLL 2017 shared task on Multilingual Parsing \cite{conll-2017-conll-2017}. For Icelandic, which is not included in this dataset, we use wikiextractor \cite{Wikiextractor2015} to get the raw Wikipedia text, and acquire lemma and POS annotations with Stanza \cite{qi2020stanza}. We hypothesize that systems trained on the Bible corpus may not generalize well to the modern language in Wikipedia. We thus additionally sample test sentences from the JW300 corpus, which is more likely to include religious language that resembles that of the bible. For JW300 we rely on the tokenization provided by the authors, but we again use Stanza for lemma and POS annotations.


For a given language and test corpus, we group gold paradigms by POS, and whether at least one form from the paradigm is attested in both training corpora. This means we have two categories for each POS: \emph{seen}, wherein at least one form is attested in both training corpora, and \emph{unseen}, wherein no forms are attested in either training corpus. We sample up to 200 paradigms from each category, ensuring that each category contributes an equal number of paradigms to the gold set. 
Then one surface form for each gold paradigm is sampled at random, in context, from the test corpus to serve as input to the systems at test time.


\section{Non-Neural Baseline for \taskabbrev}
\label{sec:baseline}
Given the set of word form clusters $c_1, ..., c_k$, where each cluster $c_i = \{f_1, ..., f_n\}$ is a collection of forms $f_j$. We start by extracting all edit trees $t = {\rm EditTree(f,f')}$ \cite{chrupala2008towards}, where $f$ and $f'$ belong to the same cluster. Let ${\rm Count}(t)$ be the count of tree $t$ across the entire training set. Further, let ${\rm MLen}(t)$ be the total number of characters which have to match in the input string, when we apply edit tree $t$. For example, for an edit tree $t$ which maps {\bf walking} to {\bf walks}, a suffix {\bf ing} must match, so ${\rm MLen}(t) = 3$. Finally, let ${\rm MStr}(t) = u$ be the string consisting of all insertions performed by the edit tree. For the given example $t$, ${\rm MStr}(t) =$ {\bf s}

When generating outputs for a given form $f$, we first form the set of all edit trees which can be applied to $f$. We then order them in the following way: $t > t'$ if ${\rm MLen}(t) > {\rm MLen}(t')$, or if the precondition lengths are equal, ${\rm Count}(t) > {\rm Count}(t')$. We then apply the top-$N$ trees to $f$ to generate all remaining forms in the inflectional paradigm of $f$. We set $N$ to the 95th percentile of paradigm sizes in our input cluster data, not counting singleton paradigms. Each slot labed is assigned based on $t$ as ${\rm MStr}(t)$. Note that this will typically not generate a slot label for the input form $f$. We, therefore, find the maximal edit tree $t$ (in the sense that it has maximal precondition length and count) which translates one of the generated forms $f'$ back into the original input form $f$. The slot label for form $f$ is then ${\rm MStr}(t)$. 

A comparison between \texttt{baseline} and our proposed POS-based system is shown in \autoref{fig:f1_by_slot_sys}. The latter outperforms \texttt{baseline} in the majority of settings, often by a large margin.

\end{document}